\documentclass[sigconf]{aamas} 

\usepackage{balance} 
\usepackage[utf8]{inputenc} 
\usepackage[T1]{fontenc}    
\usepackage{hyperref}       
\usepackage{url}            
\usepackage{booktabs}       
\usepackage{amsfonts}       
\usepackage{nicefrac}       
\usepackage{microtype}      
\usepackage{xcolor}         
\usepackage[separate-uncertainty=true]{siunitx}
\usepackage{pdflscape}

\usepackage{amsmath,amsfonts,bm}

\def\eqref#1{equation~\ref{#1}}

\def\1{\bm{1}}

\def\va{{\bm{a}}}

\def\vs{{\bm{s}}}

\DeclareMathAlphabet{\mathsfit}{\encodingdefault}{\sfdefault}{m}{sl}
\SetMathAlphabet{\mathsfit}{bold}{\encodingdefault}{\sfdefault}{bx}{n}

\DeclareMathOperator*{\argmax}{arg\,max}

\usepackage{graphicx} 
\graphicspath{{images/}}
\usepackage{amsmath} 
\usepackage{mathtools}

\usepackage{multicol}
\usepackage{placeins}
\usepackage{diagbox}
\usepackage{adjustbox}
\usepackage{verbatim}
\usepackage{cite}
\usepackage{wrapfig}
\usepackage{dcolumn}

\usepackage{enumerate}

\usepackage{algorithm} 
\usepackage[noend]{algpseudocode} 
\algnewcommand\algorithmicinput{\textbf{\textsc{Input}:} }
\algnewcommand\Input{\State \algorithmicinput}
\algnewcommand\algorithmicoutput{\textbf{\textsc{Output}:} }
\algnewcommand\Output{\State \algorithmicoutput}
\algnewcommand\algorithmicbreak{\textbf{break} }
\newcommand{\Break}{\State \algorithmicbreak}

\usepackage{subfig}
\usepackage{boldline}
\usepackage{tablefootnote}
\usepackage{multirow}

\usepackage{xcolor}

\newsavebox\CBox
\def\textBF#1{\sbox\CBox{#1}\resizebox{\wd\CBox}{\ht\CBox}{\textbf{#1}}}

\setcopyright{ifaamas}
\acmConference[AAMAS '24]{Proc.\@ of the 23rd International Conference
on Autonomous Agents and Multiagent Systems (AAMAS 2024)}{May 6 -- 10, 2024}
{Auckland, New Zealand}{N.~Alechina, V.~Dignum, M.~Dastani, J.S.~Sichman (eds.)}
\copyrightyear{2024}
\acmYear{2024}
\acmDOI{}
\acmPrice{}
\acmISBN{}

\title[Continuous Monte Carlo Graph Search]{Continuous Monte Carlo Graph Search}

\author{Kalle Kujanpää}
\authornote{Equal Contribution}
\affiliation{
  \institution{Aalto University}
  \city{Espoo}
  \country{Finland}}
\email{kalle.kujanpaa@aalto.fi}

\author{Amin Babadi}
\authornotemark[1]
\affiliation{
  \institution{Bugbear Entertainment, \\ Aalto University}
  \city{Helsinki}
  \country{Finland}}
\email{amin.babadi@bugbear.fi}

\author{Yi Zhao}
\affiliation{
  \institution{Aalto University}
  \city{Espoo}
  \country{Finland}}
\email{yi.zhao@aalto.fi}

\author{Juho Kannala}
\affiliation{
  \institution{Aalto University}
  \city{Espoo}
  \country{Finland}}
\email{juho.kannala@aalto.fi}

\author{Alexander Ilin}
\affiliation{
  \institution{Aalto University, System 2 AI}
  \city{Espoo}
  \country{Finland}}
\email{alexander.ilin@aalto.fi}

\author{Joni Pajarinen}
\affiliation{
  \institution{Aalto University}
  \city{Espoo}
  \country{Finland}}
\email{joni.pajarinen@aalto.fi}

\begin{abstract}
Online planning is crucial for high performance in many complex sequential decision-making tasks. Monte Carlo Tree Search (MCTS) employs a principled mechanism for trading off exploration for exploitation for efficient online planning, and it outperforms comparison methods in many discrete decision-making domains such as Go, Chess, and Shogi. Subsequently, extensions of MCTS to continuous domains have been developed. However, the inherent high branching factor and the resulting explosion of the search tree size are limiting the existing methods. To address this problem, we propose Continuous Monte Carlo Graph Search (CMCGS), an extension of MCTS to online planning in environments with continuous state and action spaces. CMCGS takes advantage of the insight that, during planning, sharing the same action policy between several states can yield high performance. To implement this idea, at each time step, CMCGS clusters similar states into a limited number of stochastic action bandit nodes, which produce a layered directed graph instead of an MCTS search tree. Experimental evaluation shows that CMCGS outperforms comparable planning methods in several complex continuous DeepMind Control Suite benchmarks and 2D navigation and exploration tasks with limited sample budgets. Furthermore, CMCGS can be scaled up through parallelization, and it outperforms the Cross-Entropy Method (CEM) in continuous control with learned dynamics models. 
\end{abstract}


\newcommand{\BibTeX}{\rm B\kern-.05em{\sc i\kern-.025em b}\kern-.08em\TeX}

\begin{document}


\pagestyle{fancy}
\fancyhead{}


\maketitle 


\section{Introduction}

Monte Carlo Tree Search (MCTS) is a well-known online planning algorithm for solving decision-making problems in discrete action spaces \citep{coulom2006efficient,coulom2007computing}.
MCTS achieves super-human performance in various domains such as Atari, Go, Chess, and Shogi
when a learned transition model is available \citep{schrittwieser2020mastering}. 
Therefore, recent research has tried to extend MCTS to environments with continuous state and action spaces \citep{hamalainen2014online,Rajamaeki2018,lee2020monte,kim2020monte,hubert2021learning}.

However, current MCTS approaches for continuous states and actions are limited~\citep{chaslot2008progressive,couetoux2011continuous,Rajamaeki2018,hubert2021learning}. MCTS approaches that build the search tree by discretizing the action space or otherwise limiting the growth of the tree, such as progressive widening approaches~\citep{coulom2007computing,chaslot2008progressive,couetoux2011continuous}, do not scale up well to high-dimensional action spaces and complex problems due to the search space increasing
exponentially with the planning horizon and action dimensionality \citep{browne2012survey,yee2016monte}.
To alleviate this problem, learning-based approaches~\citep{Rajamaeki2018,hubert2021learning}
reduce the required planning horizon when a sufficiently large number of training samples is available but do not solve the underlying
problem of increasing search space size.

To solve the search space explosion problem in MCTS-based continuous planning, this paper presents Continuous Monte Carlo Graph Search (CMCGS), an
extension of MCTS to the continuous control problem. Similarly to MCTS, CMCGS employs an iterative mechanism for building a search graph that yields the next action to execute.
At each iteration, a series of operators is used to grow the search graph and update the information stored in the graph nodes. CMCGS uses state clustering and Gaussian action bandits to deal with the challenges posed by high-dimensional continuous states and actions and limit the search space explosion.

Our experiments show that CMCGS outperforms strong MCTS and other planning baselines in many challenging benchmarks given limited interaction with the environment, including high-dimensional environments from the DeepMind Control Suite (DMC) \citep{tunyasuvunakool2020}. We also show that CMCGS can be efficiently parallelized to scale up and demonstrate its generality and robustness with learned dynamics models in high-dimensional visual tasks, where CMCGS outperforms the Cross-Entropy Method (CEM) \citep{rubinstein2004cross, weinstein2013open}. Moreover, CMCGS shows strong performance by being superior to CEM in environments that require complex exploration, and our algorithm also performs well in difficult, high-dimensional environments with informative rewards, where MCTS-based methods fail.

\begin{figure*}[ht]
  \centering
  \includegraphics[width=1\linewidth]{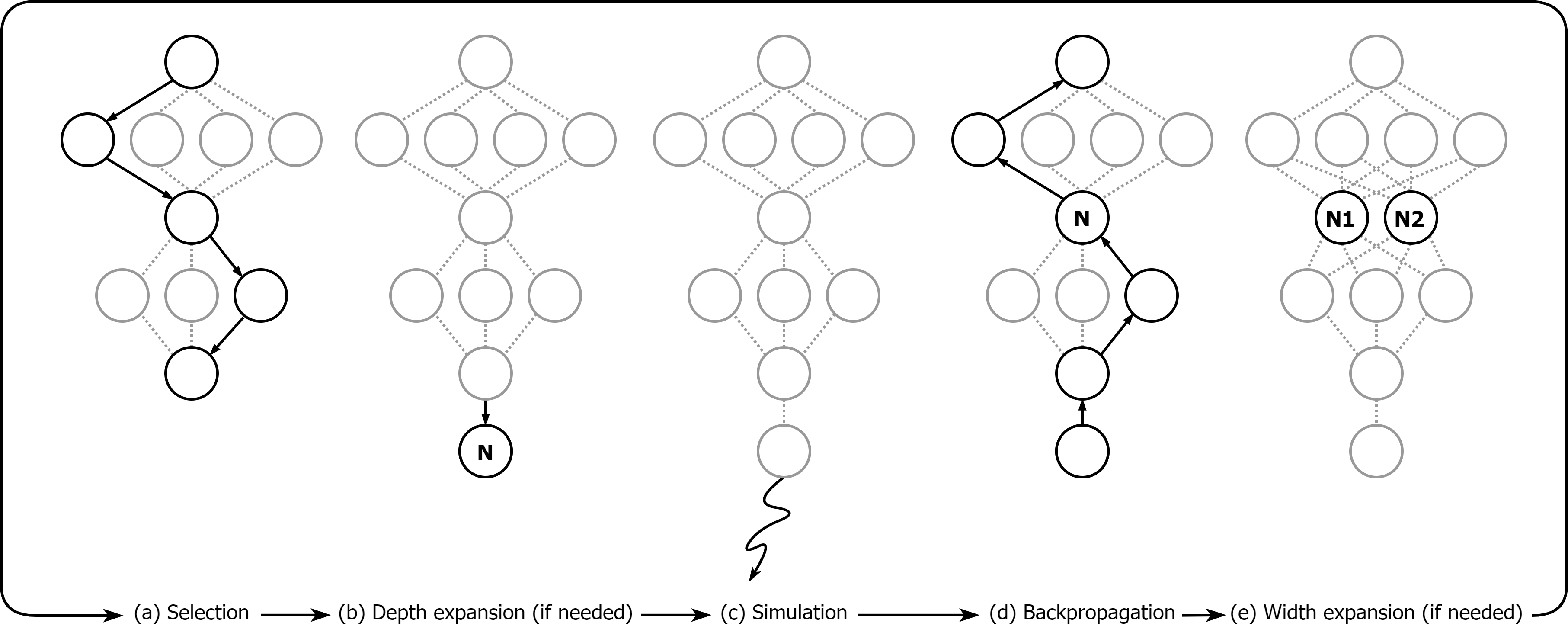} 
  \caption{Core steps in one iteration of Continuous Monte Carlo Graph Search (CMCGS). a) Starting from the root node, the graph is navigated via action sampling and node selection until a sink node is reached. b) If there is enough experience collected in the final layer of the graph and the maximum depth has not been reached, a new layer containing a new node \textbf{N} is initialized.
  c) A trajectory of random actions is simulated from the graph's sink node to approximate the value of the state. d) The computed accumulated reward is backed up through the selected nodes, updating their replay memories, policies, and state distributions. e) If a new cluster of experience data is found in a previous layer of the graph, all nodes in that layer are updated based on the new clustering information (in this example, the node \textbf{N} is split into two new nodes \textbf{N1} and \textbf{N2}).}
  \label{fig:mcgs_diagram}
\end{figure*}

\section{Related Work}
\textbf{Monte Carlo Tree Search} (MCTS), a decision-time planning method in discrete action spaces~\citep{coulom2006efficient,kocsis2006bandit,browne2012survey}, is a core component of the success of the computer Go game \citep{gelly2011monte,silver2017mastering,silver2018general}. MCTS also shows advantages in general game playing \citep{finnsson2008simulation,guo2014deep,anthony2017thinking,grill2020monte}. Several recent works combine MCTS with a learned dynamic model lifting the requirement for access to a simulator~\citep{schrittwieser2020mastering,ye2021mastering}. Grouping similar or equivalent states reached via different trajectories in MCTS has been proposed \citep{childs2008transpositions}, but recent methods mostly focus on the discrete state setting \citep{leurent2020monte, czech2021improving}, which differs from our continuous setup.

\textbf{Online planning in continuous action spaces}. In continuous action spaces, there can be an infinite number of actions, which significantly increases the size of the MCTS search tree and makes applying out-of-the-box MCTS infeasible. Instead, CEM is commonly used in the continuous action domain as an online planning method \citep{rubinstein1997optimization,rubinstein2004cross, weinstein2013open,chua2018deep,hafner2019learning}. Many methods combine CEM with a value function or policy in different ways to improve its performance \citep{negenborn2005learning,lowrey2018plan,bhardwaj2020information,hatch2021value,hansen2022temporal}. The main limitation of CEM compared to MCTS (and this work) is that CEM models the whole action trajectory using a single sampling distribution. This could be translated into the context of MCTS as using only one node at each layer of the search tree, and it can limit the exploration capabilities of CEM in environments where there are several ways of controlling the agent, for example, going around an obstacle using two different ways. Covariance Matrix Adaptation Evolution Strategy (CMA-ES) is an evolutionary gradient-free algorithm for black-box continuous optimization \citep{hansen1996adapting, hansen2003reducing}. Although it has not seen much use in continuous control due to its computational complexity and memory requirements, applying it as an alternative to CEM has been proposed \citep{heidrich2009uncertainty, naderi2017discovering, duan2016benchmarking}.

\textbf{MCTS in continuous action spaces}
Adopting MCTS in continuous action spaces requires mitigating the requirement of enumerating all actions as in the discrete case. Several works use progressive widening to increase the number of child actions of a node as a function of its visitation frequency \citep{coulom2007computing,chaslot2008progressive,couetoux2011continuous,moerland2018a0c}. The action space can be split into bins by factorizing across action dimensions \citep{hamrick2020role, tang2020discretizing}. 
A line of work extends MCTS to continuous action spaces via Hierarchical Optimistic Optimization (HOO) \citep{bubeck2008online,munos2014bandits,mao2020poly,quinteiro2021limited}. HOO hierarchically partitions the action space by building a binary tree to incrementally split the action space into smaller subspaces. Voronoi Optimistic Optimization (VOO) builds on similar ideas as HOO. It uses Voronoi partitioning to sample actions efficiently and is guaranteed to converge \citep{kim2020monte}. VG-UCT relies on action value gradients to perform local policy improvement \citep{lee2020monte}. Furthermore, several works grow the search tree based on sampling. Kernel regression can be used to generalize the value estimation and, thus, limit the number of actions sampled per node \citep{yee2016monte}. However, our experiments indicate that despite the presented strategies, these methods still struggle with the rapidly growing search space, especially with high-dimensional states and actions. We propose tackling this challenge by representing the search space as a graph instead of a tree, clustering similar states into search nodes, and using Gaussian action bandits for the nodes. In practice, CMCGS significantly improves on the continuous MCTS methods, especially with limited search budgets.

\textbf{State abstractions in MCTS} Our approach to clustering similar states into stochastic action bandit nodes is closely related to state abstraction and aggregation methods. State abstraction helps mitigate the impact of the combinatorial explosion in planning \citep{abel2016near}. In state aggregation, many states are combined into abstract ones \citep{li2006towards, van2006performance} and it has been proposed for handling the high branching factor in stochastic environments \citep{hostetler2014state}. Hierarchical MCTS through action abstraction is closely related \citep{bai2016markovian}. Progressive Abstraction Refinement in Sparse Sampling (PARSS) builds and refines state abstractions as the search progresses \citep{hostetler2017sample}. However, these methods, including PARSS, generally assume a discrete action space, incompatible with our setup. State abstractions can also be used in a continuous setting when combined with progressive widening \citep{sokota2021monte}. Our results indicate that relying on progressive widening makes the proposed method uncompetitive with CMCGS. Elastic MCTS uses state abstractions for strategy game playing, but lacks a progressive widening or discretization strategy to handle the high-dimensional continuous action spaces prevalent in our experiments \citep{xu2023elastic}.

There are also learning-based approaches to online planning that represent the policy using a neural network and sample from it when expanding the tree \citep{ahmad2020marginal,hubert2021learning}. By leveraging the learned policy, these methods achieve promising performance with limited search budgets. However, we focus only on the core search component and leave comparisons with learning-based methods to future work.

\section{Preliminaries}

Planning and search is arguably the most classic approach for optimal control \citep{fikes1971strips, mordatch2012discovery, tassa2012synthesis}. The idea is to start with a (rough) estimation of the action trajectory and gradually improve it through model-based simulation of the environment. In the case of environments with complex dynamics, this is usually done online by optimizing only a small part of the action trajectory at each step. Then, the first action of the optimized trajectory is executed. After that, the whole process is repeated, starting from the newly visited state. This process is also known as closed-loop control, or model-predictive control (MPC). MPC is an effective approach for optimal control in complex real-time environments \citep{samothrakis2014rolling, gaina2017population, babadi2018intelligent}.

Monte Carlo Tree Search (MCTS)~\citep{coulom2006efficient} is one of the most popular MPC algorithms in environments such as video games \citep{perez2014knowledge,holmgaard2018automated} that require principled exploration.
MCTS starts by initializing a single-node search tree using the current state $s_t$ as the root and grows the tree in an iterative manner, where one node is added to the tree at each iteration.
The iterative process of MCTS is comprised of the following four key steps \citep{browne2012survey}:

\begin{enumerate}
    \item \textbf{Selection (Tree Policy)}: In this step, the algorithm selects one of the tree nodes to be expanded next. This is done by starting from the root and navigating to a child node until a node with at least one unexpanded child is visited using a selection criterion, whose goal is to balance the exploration-exploitation trade-off.
    \item \textbf{Expansion}: In this step, MCTS expands the search tree by adding a (randomly chosen) unvisited child to the selected node from the previous step.
    \item \textbf{Simulation (Default Policy)}: After expanding the selected node, the newly added node is evaluated using a trajectory of random actions and computing the return.
    \item \textbf{Backup}: In the final step, the computed return is backed up through the navigated nodes, updating the statistics stored in each node.
\end{enumerate}

For a more detailed description of the MCTS algorithm, the reader is referred to \citet{browne2012survey}.

\section{Continuous Monte Carlo Graph Search}

\label{sec:method}

We propose an algorithm for continuous control called Continuous Monte Carlo Graph Search (CMCGS). We assume a discrete-time, discounted, and fully-observable Markov Decision Process (MDP) $M = \langle \mathcal{S}, \mathcal{A}, P, R, \gamma \rangle$ with continuous states and actions. $\mathcal{S}$ is the state space, $\mathcal{A}$ the action space, $P$ the transition probability function, $R$ the reward function, and $\gamma \in [0, 1]$ the discount factor. The objective of CMCGS is to find the action $a \in \mathcal{A}$ that maximizes the return in $M$ by performing closed-loop control with model-based simulation of the environment. 
CMCGS is an iterative online planning algorithm in which each iteration contains the same four steps as MCTS with an additional width expansion step (see Fig.~\ref{fig:mcgs_diagram}). Instead of building a search tree, CMCGS performs a search on a layered graph in which each layer corresponds to one step of the underlying MDP.
We assume that we have access to a dynamics model $p(\vs_{t+1}, r_{t+1} | \vs_t, \va_t)$ that can either be an exact model of the environment or a learned approximation.

The core idea of CMCGS is to use state clustering, Gaussian action bandits, and a layered directed search graph to tackle the challenges posed by search state explosion and high-dimensional continuous states and actions. Each node $q$ in the $t$-th layer of the CMCGS graph corresponds to \emph{a cluster of states visited at timestep $t$}. The node is characterized by its state distribution $p_q(\vs_t)$ and policy $\pi_q(\va_t)$, which describe the states assigned to this cluster and the preferred actions taken from this cluster during planning.
Both distributions $p_q(\vs_t)$ and $\pi_q(\va_t)$ are modeled as Gaussian distributions with diagonal covariance matrices. The policy $\pi_q(\va_t)$ is initialized as a Gaussian $\mathcal{N}(\mu = \frac{1}{2} (\va_{\min} + \va_{\max}), \sigma = \frac{1}{2} (\va_{\max} - \va_{\min}))$. During planning, the distributions are fit to the data in the replay buffer. The replay buffer contains 
tuples $e = (\vs_t, \va_t, \vs_{t+1}, R, q)$, where $q$ is the node to which state $\vs_t$ belongs and $R$ is the observed trajectory return. 
We denote by $\mathcal{D}_q$ all the tuples in the replay buffer that correspond to node $q$ and by $Q_t$ the set of nodes in layer $t$.

The search graph is initialized with $d_\text{init}$ layers and one node per layer. Initializing the graph with multiple layers improves the early search efficiency. The CMCGS algorithm then iterates the following steps (see Fig.~\ref{fig:mcgs_diagram} for a graphical illustration).

\begin{enumerate}[(a)]
\item \textbf{Selection}: The selection mechanism is repeatedly applied to navigate through the search graph starting from the root node until a sink node (a node without children) or a terminal state of the MDP is reached (Fig.~\ref{fig:mcgs_diagram}a).
When navigating through the graph, we use an epsilon-greedy-inspired policy: with probability $\epsilon$, we sample an action from the node policy, the Gaussian distribution $\pi_q(\va)$, and with probability $1-\epsilon$, we uniformly sample one of the top actions in $\mathcal{D}_q$ (greedy action) and modify it with a small Gaussian noise for local improvement\footnote{This approach enables the agent to leverage past experience efficiently while exploring new options, leading to better performance on complex tasks.}: $\epsilon_\text{top} \sim \mathcal{N}(0, \mathcal{E}_\text{top} \cdot  (\va_{\max} - \va_{\min}))$.
The current state $\vs_t$ is updated to $\vs_{t+1}$ according to the dynamics model $p(\vs_{t+1}, r_{t+1} | \vs_t, \va_t)$, and the node in the next layer is selected by maximizing the probability of the updated state according to the node state distributions:
\[
  q \gets \argmax\limits_{q \in Q_{t+1}} p_q(\vs_{t+1})
\]

\item \textbf{Depth expansion}: 
If the last layer of the graph has collected enough experience (the number of samples is greater than threshold $m$) and the maximum graph depth has not been reached, we add a new layer to the graph (Fig.~\ref{fig:mcgs_diagram}b). The new layer contains a single node initially. Without the threshold $m$, the depth of the search graph would grow too fast.

\item \textbf{Simulation}: Similar to MCTS, we simulate a random trajectory using random actions starting from the sink node and compute the trajectory return $R$ (Fig.~\ref{fig:mcgs_diagram}c). The length of the rollout is controlled by a hyperparameter $N_r$, but the rollout is interrupted if a terminal state is encountered.

\item \textbf{Backup}: We traverse the search graph backward and update the 
node distributions $p_q(\vs)$ and $\pi_q(\va)$ for all the nodes visited in the selection procedure (Fig.~\ref{fig:mcgs_diagram}d). The replay buffers $\mathcal{D}_q$ of the visited nodes are updated. For each affected node $q$, the Gaussian state distribution $p_q(\vs)$ is fitted to the states $\vs$ in the node replay buffer $\mathcal{D}_q$ such that its mean is equal to the state mean and standard deviation to the state standard deviation.
For the action distributions $\pi_q(\va)$, we use a threshold-based strategy to encourage sufficient exploration: the node policy $\pi_q(\va)$ is updated only if the replay buffer of the node $q$ contains enough experience ($|\mathcal{D}_q| > m / 2$). The dimension means of $\pi_q(\va)$ are fitted to the highest-scoring (elite) experiences in $\mathcal{D}_q$, similarly to CEM. To update the dimension variances of $\pi_q(\va)$, we adopt the Bayes' rule by defining a conjugate prior (inverse gamma) on the variances, compute the posteriors using the elite samples, and select the posterior means as the variances. This prevents the variance from decaying too fast. Formally, we assume that the dimension variances $\sigma^2$ of $\pi_q(\va)$ follow the inverse gamma distribution with prior hyperparameters $\alpha$ and $\beta$. Given $n$ elite samples $(\va_{t,1}, \dots, \va_{t,n})$ and the new dimension means $\mu$ of $\pi_q(\va)$, the posterior hyperparameters of the inverse gamma distribution are:
\begin{align*}
    \alpha_\text{posterior} = \alpha_\text{prior} + \frac{n}{2},\text{ }
    \beta_\text{posterior} = \beta_\text{prior} + \frac{\sum_{i=1}^n (\va_{t,i} - \mu)^2}{2}
\end{align*}
and the new dimension variances $\sigma^2$ of $\pi_q(\va)$ are equal to the means of the inverse gamma posterior distributions:
\begin{equation}
\sigma^2 = \frac{\beta_\text{posterior}}{\alpha_\text{posterior} - 1}.
\label{eq:bayes}
\end{equation}

\item \textbf{Width expansion}: 
To prevent the rapid search graph growth and maintain efficiency, CMCGS computes the maximum desired number
$C_t$ of nodes in layer $t$ using a heuristic rule
\[
C_t = \min \left ( n_{\max}, \left \lfloor n_t / m \right \rfloor \right )
\]
where $n_t$ is the number of transitions collected at timestep $t$ and stored in the replay buffer $\mathcal{D}$, $m$ the threshold hyperparameter, and $n_\text{max}$ the hard limit on the number of nodes.
If $|Q_t|$, the number of nodes in layer $t$, is smaller than $C_t$, we attempt to cluster the states visited at timestep $t$ into $|Q_t| + 1$ clusters. We use Agglomerative Clustering with Ward linkage in our experiments due to its fast running time with the amount of data needed by CMCGS 
\citep{pedregosa2011scikit, ward1963hierarchical}. Using Euclidean distance as a metric is feasible both in low-dimensional state spaces and learned latent spaces.
The clustering is approved if each cluster contains at least $m / 2$ states to prevent degenerate clusters that are unlikely to be selected for states encountered in the future. Then, a new node is added to the layer (Fig.~\ref{fig:mcgs_diagram}e), all experience tuples in the replay buffer are re-assigned to the corresponding nodes and the state and action distributions $p_q(\vs)$ and $\pi_q(\va)$ of the affected nodes are updated. Otherwise, no new node is created, and clustering is attempted again when the layer has collected $m/2$ new samples to limit the number of calls to the clustering algorithm and, therefore, speed up the algorithm.

\end{enumerate}

\begin{algorithm}[!t]\caption{Continuous Monte Carlo Graph Search}
\label{alg:mcgs}
\begin{algorithmic}[1]
\Function{CMCGS}{$\vs_0$}
	\State $q_\text{root} \gets$ Node($s_0$, parent=None)
    \State Initialize search graph with $d_\text{init}$ layers and $q_\text{root}$ as the root.
    \While{within computational budget}
        \State $\tau, R, \vs_d \gets \Call{GraphPolicy}{\vs_0,q_\text{root}}$
        \If{$\vs_d$ is not terminal}
        \State $R \gets R + \Call{RandomRollout}{\vs_d}$
        \EndIf
        \State $\Call{Backup}{\tau, R}$
    \EndWhile
    \State \Return $\Call{GetBestAction}{q_\text{root}}$
\EndFunction

\Function{GraphPolicy}{$\vs, q$}
	\State $\tau \gets \emptyset$
	\State $R \gets 0$
    \State $d \gets 1$
    \State $\phi \gets \begin{cases}
        1\text{, if  }\mathcal{U}(0, 1) < \epsilon \\
        0\text{, otherwise}
    \end{cases}$
    \While{$\vs$ is not terminal \textbf{and} $d \le N_\text{layers}$}
    	\State $\va \sim \phi \pi_q(\va) + (1-\phi) \pi_{q,\text{greedy}}(\va)$ \State Apply action $\va$, observe new state $\vs'$, reward $r$
    	\State $\tau \gets \tau \cup \lbrace \langle \vs,\va,\vs',q \rangle \rbrace$
    	\State $R \gets R + r$
    	\If{$\vs'$ is not terminal \textbf{and} $d = N_\text{layers}$}
			\State Try adding a new layer with one node to the graph.
		\EndIf
		\If{$s'$ is terminal \textbf{or} $d = N_\text{layers}$}
			\Break
		\EndIf
    	\State $q \gets \argmax\limits_{q \in Q_{t+1}}
    p_q\left(\vs'\right)$
    	\State $\vs \gets \vs'$
        \State $d \gets d + 1$
    \EndWhile
    \State \Return $\tau,R,\vs$
\EndFunction

\Function{Backup}{$\tau, R$}
	\For{\textbf{each} $\langle \vs_t,\va_t,\vs_{t+1},q_t \rangle \in \tau$}
	\State $\Call{BufferStore}{\langle \vs_t,\va_t,R,\vs_{t+1},q_t \rangle}$
	\State Compute $C_t$, the desired number of clusters in layer $t$
    \If{$|Q_t| < C_t$}
	\State Cluster states $\{\vs_t\}$ in layer $t$ into $|Q_t| + 1$ clusters
	\If{clustering was successful}
	\State Replace nodes in layer $t$ with the new nodes
    \For{$q \in Q_t$}
    \State Estimate state distribution $p_q(\vs)$ 
    \State Estimate policy $\pi_q(\va)$ 
    \EndFor
	\EndIf
	\EndIf
	\If{no new cluster found in layer $t$}
	\State Update $p_q(\vs)$ and $\pi_q(\va)$ of the visited node $q_t$ 
	\EndIf
	\EndFor
\EndFunction
\end{algorithmic}
\end{algorithm}

When the simulation budget has been exhausted, the agent's next action is determined by the experience stored at the root node. If the search has access to a deterministic environment simulator, the first action of the trajectory with the highest return is chosen. If searching with a learned dynamics model or the environment is stochastic, the average of the top actions in the replay memory of the root node is returned to prevent the exploitation of model inaccuracies or individual samples. Each of the five steps of CMCGS can be performed efficiently using batch operations, making it possible to collect multiple trajectories in parallel.
The pseudocode of the CMCGS algorithm is shown in Algorithm \ref{alg:mcgs}.

\section{Experiments}\label{sec:experiments}

We run four sets of experiments to evaluate the performance, generality, and robustness of CMCGS. First, we present a motivating example where CEM fails due to its inability to explore properly in an environment where a multimodal representation of the action distribution is useful. Second, we evaluate CMCGS in continuous control tasks with limited sample budgets to demonstrate that it outperforms relevant baselines in a variety of tasks with different levels of exploration required. Third, we demonstrate that CMCGS can be parallelized and can successfully utilize learned models for large-scale planning in complex tasks by showing that it achieves significantly better planning performance than CEM in image-based continuous control environments. Finally, we perform extensive ablation studies to better understand the design choices of CMCGS.

\subsection{Toy Environment}

\begin{figure}[t]
\centering
\begin{minipage}{.325\linewidth}
\centering
\includegraphics[width=\linewidth,trim={15mm 15mm 15mm 15mm},clip]{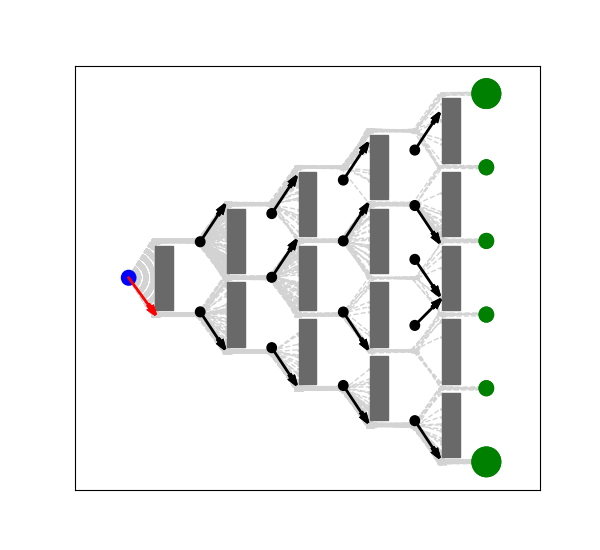}
CMCGS (Success)
\end{minipage}
\hfil
\begin{minipage}{.325\linewidth}
\centering
\includegraphics[width=\linewidth,trim={15mm 15mm 15mm 15mm},clip]
{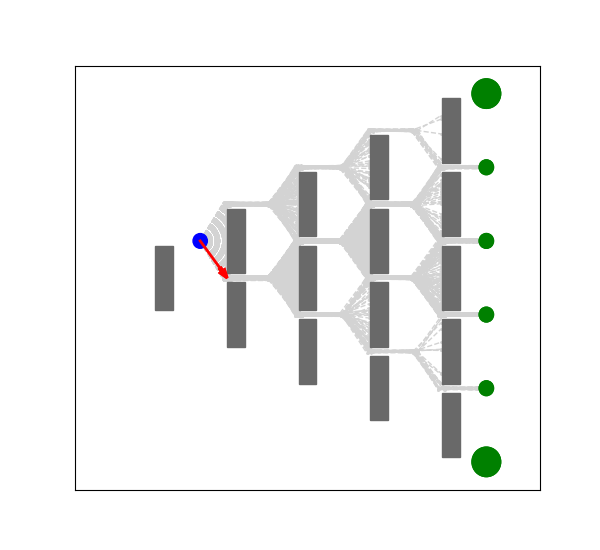} \\
CEM (Partial Failure)
\end{minipage}
\hfil
\begin{minipage}{.325\linewidth}
\centering
\includegraphics[width=\linewidth,trim={15mm 15mm 15mm 15mm},clip]
{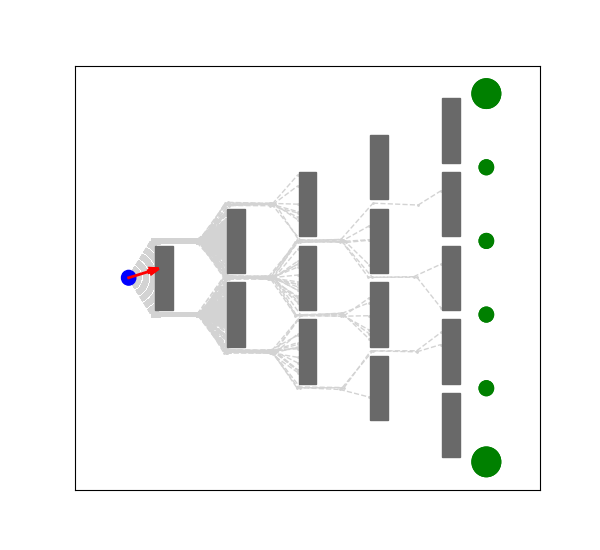} \\
CEM (Failure)
\end{minipage}
\caption{An illustration of exploration in the toy environment. For clarity, the actions have been clipped to be in the range $[-1, 1]$. The agent, the chosen action, the explored trajectories, and the different-sized rewards are represented by the blue dot, red arrow, grey dashed lines, and green dots, respectively. The search nodes of CMCGS and the corresponding state and action means are illustrated with black dots and arrows. The left image illustrates how CMCGS explores with state-dependent policies. In the other pictures, we see how CEM fails in the environment. CEM can either fail to discover the large reward and choose a suboptimal action (center) or completely fail to handle the multimodality required by the environment (right).}
\label{fig:toy_env_illustration}
\end{figure}

We design a toy environment where the agent samples $N$ actions $a_1, \dots, a_N$ sequentially and receives a basic reward of $r = 0.5$ if the absolute value of each action is greater than $1$. An additional reward of $0.5$ is given if the sign of all actions is the same. The reward is formalized as
\begin{align*}
    r = \begin{cases} 0.5, & \text{if } \forall i \in \{ 1, \dots, N \}: |a_i| > 1 \\
    1, & \text{if } \forall i \in \{1, \dots, N\} : |a_i| > 1 \, \land  \\
    & \hspace{8pt} \forall i \in \{2, \dots, N\} : \text{ sgn}(a_i) = \text{ sgn}(a_1) \\
\end{cases}
\end{align*}
All action bandits are initialized to $\mathcal{N}(0, 1)$ to eliminate the effect of the initial action distribution on the performance, the action space is equal to $\mathbb{R}$, and the state equal to the y-coordinate of the agent (see Fig.~\ref{fig:toy_env_illustration}). Finding the basic reward of $r = 0.5$ 
is easy, but performing principled exploration to find the additional component is a challenge. Furthermore, being able to represent complex multimodal action distributions is a valuable asset in this environment. We let $N=5$ and compare our method to random shooting and CEM. The results are plotted in Table~\ref{tab:toyenv} and the agents' behavior is illustrated in Fig.~\ref{fig:toy_env_illustration}. The results show that CEM struggles to explore the environment properly, and CMCGS outperforms random shooting due to directed exploration. 
CEM sometimes fails to receive the basic reward of $0.5$ due to the inability to represent the multimodal action distribution natural to this problem.

\begin{table}[t]
\caption{The success rates of achieving the small and large rewards and the average rewards achieved by different algorithms in the toy environment plus-minus two standard errors. 
}
\centering
\begin{tabular}{lccc}
\toprule
Method & \,$r\ge0.5$\, & \,$r=1.0$\, & Average Reward \\
\midrule
CMCGS & $\textBF{1.00}$ & $\textBF{0.99}$ & $\textBF{0.995} \pm 0.002 $ \\
Random shooting & $\textBF{1.00}$ & $0.89$ & $0.943 \pm 0.010 $ \\
CEM & 0.74 & 0.57 & $0.655 \pm 0.028 $ \\
\bottomrule
\end{tabular}
\label{tab:toyenv}
\end{table}

\subsection{Continuous Control with Limited Interaction}

\begin{figure}[tp]
\centering
\subfloat[2D navigation task using circular obstacles (\textit{2d-navigation-circles}).]
{\includegraphics[width=0.75\linewidth,trim={40mm 43mm 35mm 42mm},clip]{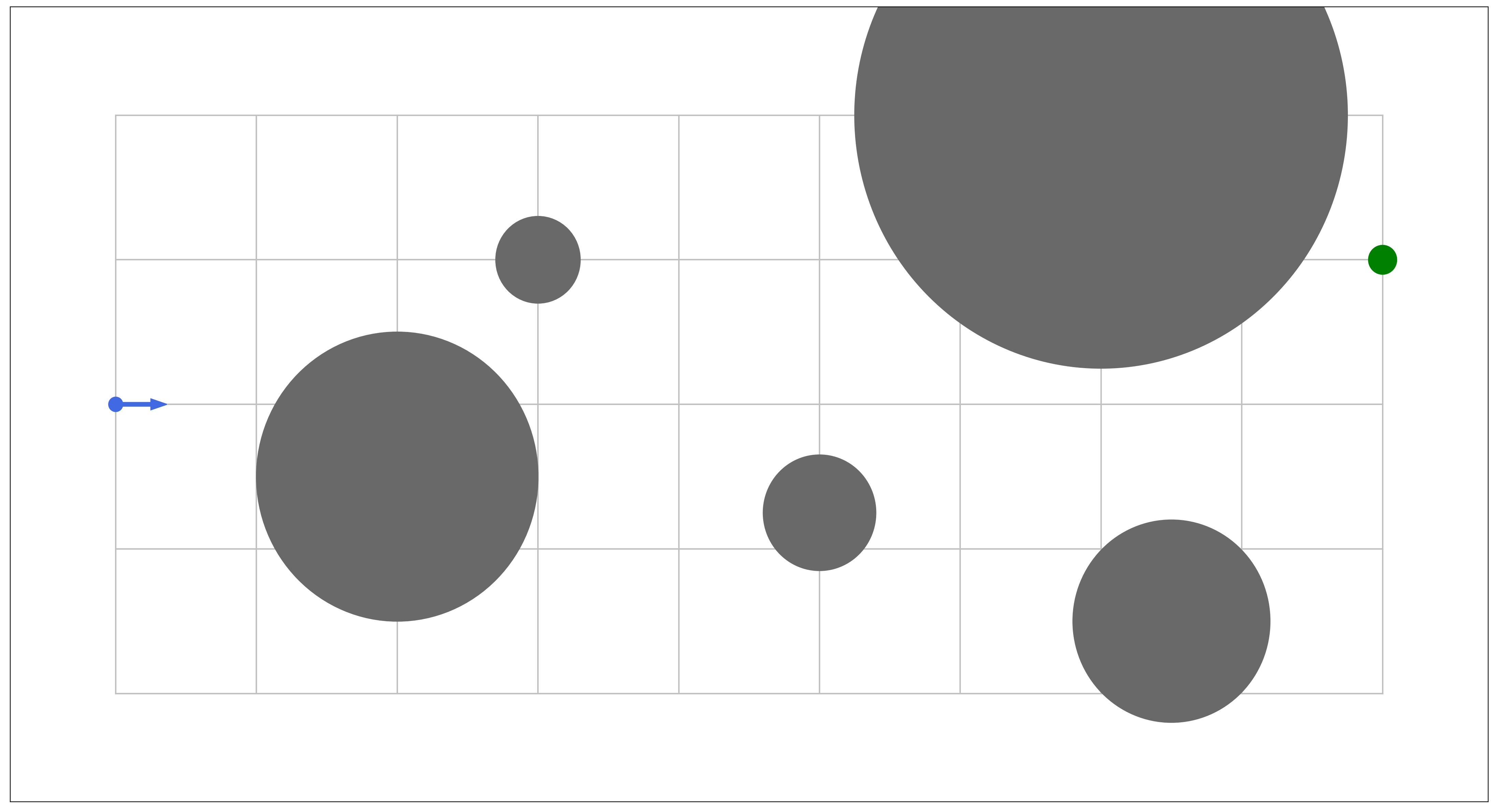}
\label{fig:2d-envs-circle}}
\\
\subfloat[2D reacher task using $15$-linked arm (\textit{2d-reacher-fifteen-poles}).]
{\includegraphics[width=0.75\linewidth,trim={35mm 15mm 30mm 20mm},clip]{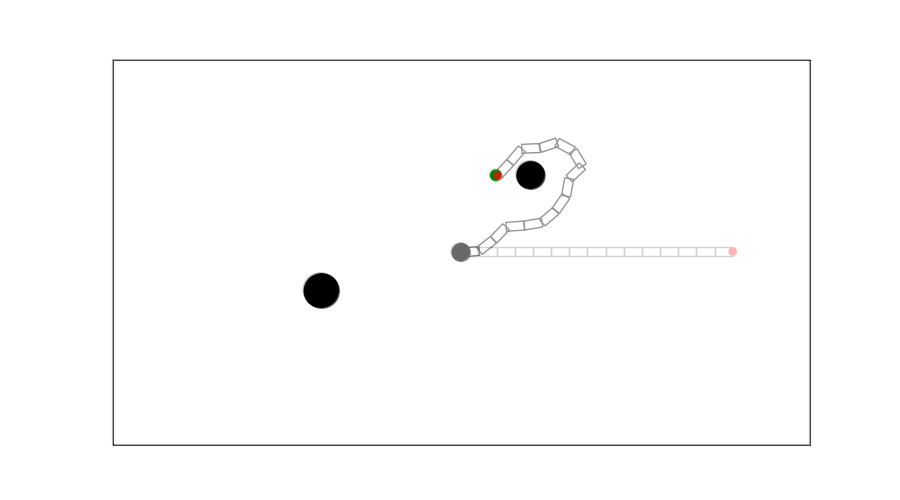}
\label{fig:2d-reacher-fifteen}}
\caption{Custom environments used in the experiments.}
\label{fig:2d-envs}
\end{figure}

Second, we compare CMCGS to the strong planning algorithms (1)~Monte Carlo Tree Search with Progressive Widening \citep[MCTS-PW]{chaslot2008progressive,couetoux2011continuous,coulom2007computing}, (2) Voronoi Optimistic Optimization Applied to Trees \citep[VOOT]{kim2020monte}, (3) MCTS with Iteratively Refined State Abstractions (MCTS-ABS) \citep[MCTS-ABS]{sokota2021monte}, (4) Cross-Entropy Method \citep[CEM]{rubinstein1997optimization}, and (5) Covariance Matrix Adaptation Evolution Strategy \citep[CMA-ES]{hansen2003reducing} to demonstrate the power of CMCGS in continuous control tasks when the number of environment interactions is a bottleneck. This can be the case in, for instance, real-life robotics or when a computationally very intensive and accurate simulator is used. In these tasks, we evaluate the planning performance of the algorithms on real dynamics. We also include standard random shooting (RS) as a baseline. More details about the baselines and their implementations can be found in Appendix~\ref{app:baselines}.

The environments we use in these experiments include:

\begin{enumerate}
    \item \textbf{2D Navigation}: We developed an environment for the control of a particle that should reach the goal (shown in green in Fig.~\ref{fig:2d-envs-circle}) without colliding with the obstacles. We used two variations of this environment, one with circular and another with rectangular obstacles. In these environments, a new action is applied whenever the particle reaches a vertical gray line. The main challenge in these environments is that the agent has to explore to find the green reward and avoid colliding with the obstacles. Full environment specifications can be found in Appendix~\ref{app:custom_envs}.
    \item \textbf{2D Reacher}: Since \textit{2D Navigation} environments only have $1$ Degree of Freedom (DOF), we also developed a variation of the classic reacher task for a 2D multi-link arm. In this environment, only a sparse binary reward is used for reaching the goal (shown in green in Fig.~\ref{fig:2d-reacher-fifteen}), and there are obstacles (shown in black) that block the path. We use $15$- and $30$-linked arms to evaluate the scalability of CMCGS to high-dimensional action spaces. The 15-link variation is shown in Fig.~\ref{fig:2d-reacher-fifteen}. Full environment specifications can be found in Appendix~\ref{app:custom_envs}.
    \item \textbf{DeepMind Control Suite}: We also used several popular environments provided by DeepMind Control Suite \citep{tassa2018deepmind} in our experiments. These environments pose a wide range of challenges for continuous control with low- and high-dimensional states and actions. In this part of the evaluation, we use proprioceptive observations.
\end{enumerate}

We evaluate CMCGS and the baselines CEM,  MCTS-PW, VOOT, MCTS-ABS, CMA-ES, and random shooting using different simulation budgets per control timestep. For each budget, we used $400$ random seeds for 2D navigation tasks, $200$ for 2D reacher tasks, and $100$ for DMC tasks. All experiments used the same set of hyperparameters (see Appendix~\ref{app:hyperparams}) that were carefully tuned to maximize the performance on the chosen environments.

\begin{figure}
  \begin{center}
  \includegraphics[width=\linewidth]{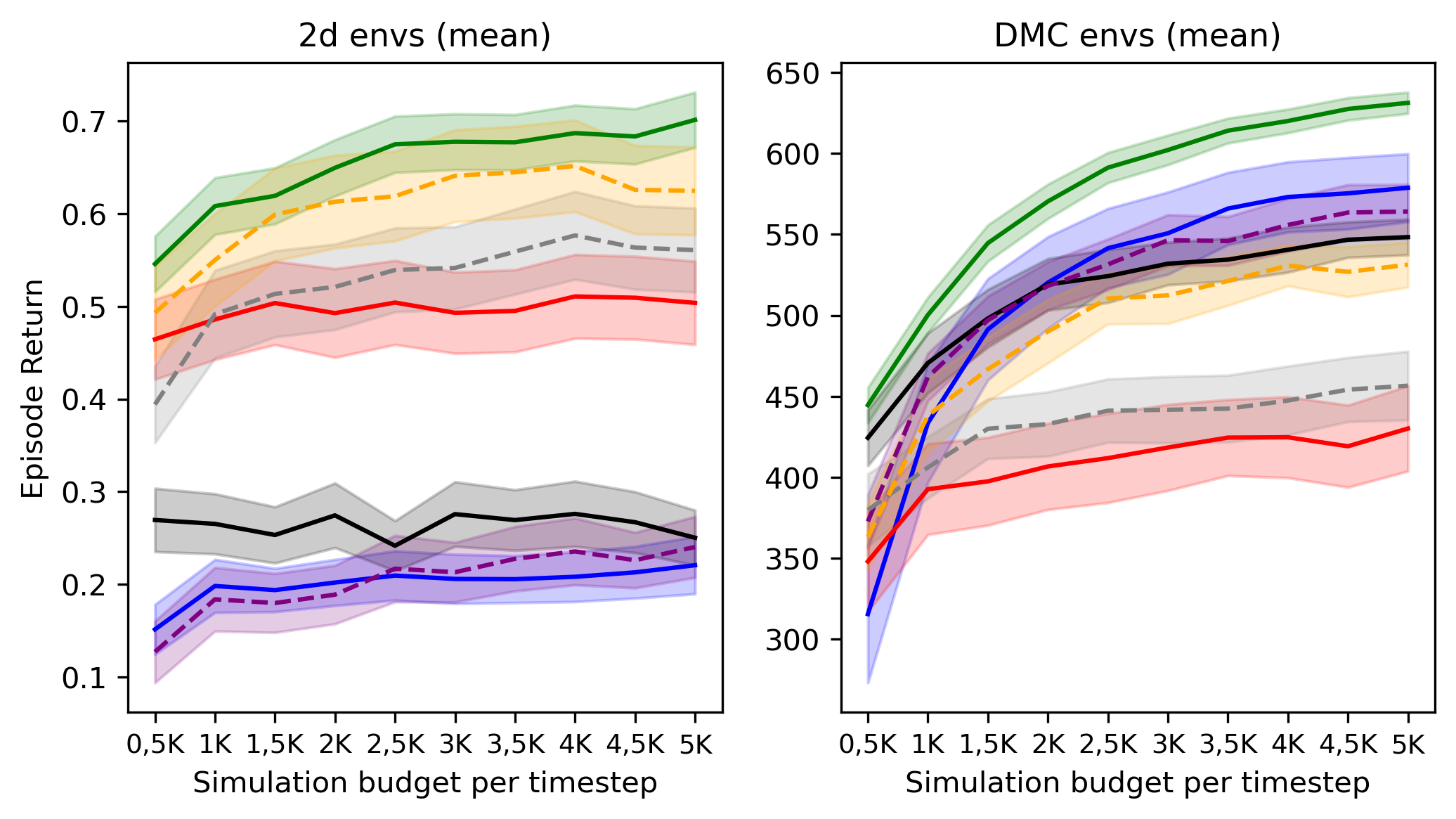}
  \includegraphics[width=\linewidth,trim={2mm 2mm 2mm 2mm},clip]{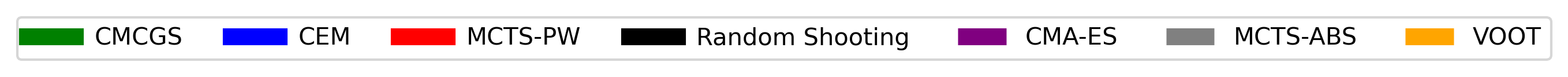}
  \end{center}
  \caption{The mean episode returns averaged over the custom 2D environments that test exploration (left) and the environments from DeepMind Control Suite (right).
  }
  \label{fig:reward_summary}
\end{figure}

\begin{table}
\end{table}

\begin{figure*}[t]
\centering
{{\includegraphics[width=1.0\linewidth]{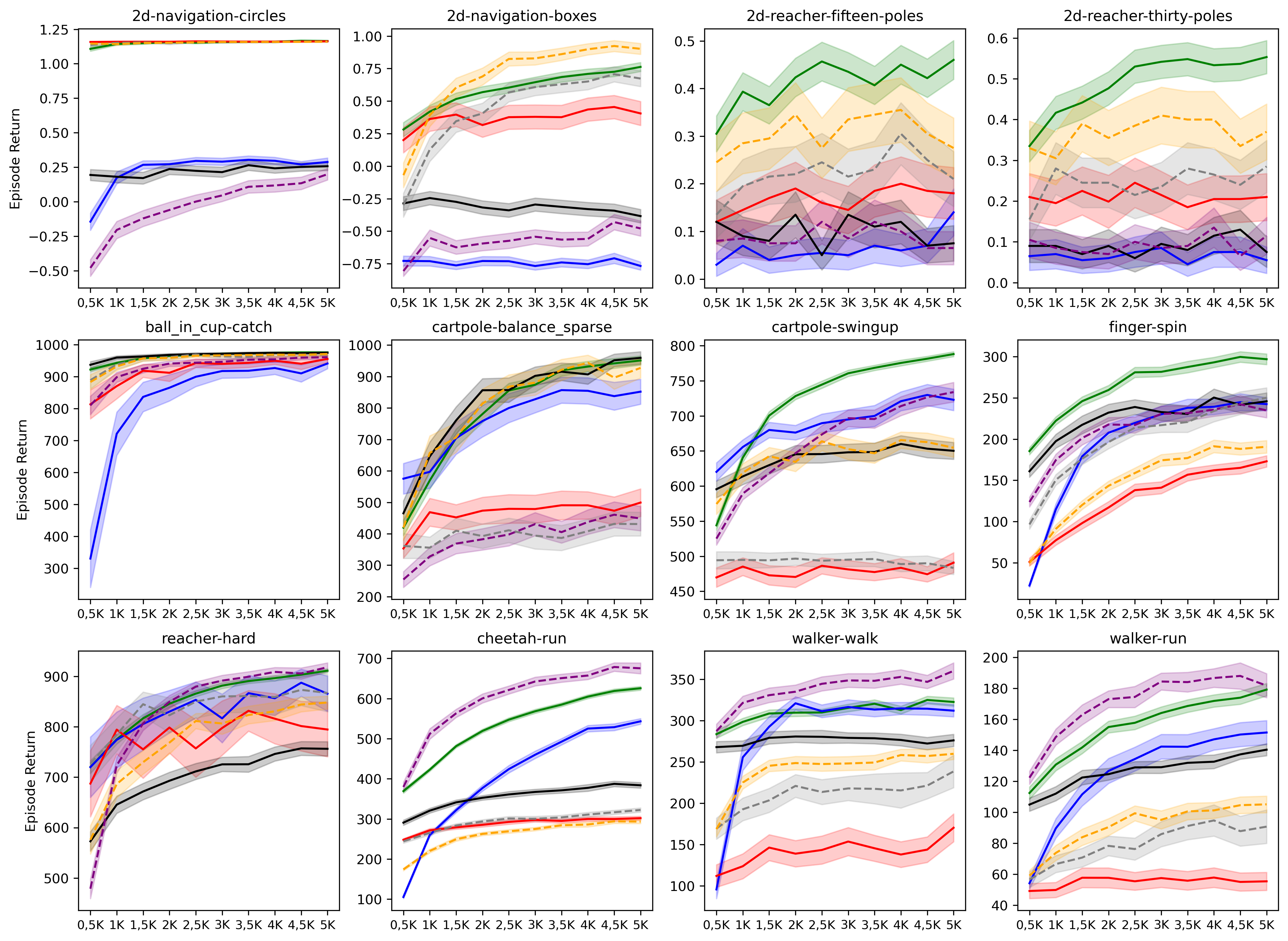} }}
\vfill
{{\includegraphics[width=0.8\linewidth,trim={2mm 2mm 2mm 2mm},clip]{varying_budget-leg} }}
\caption{Reward plots for different simulation budgets per timestep.
\textit{2d-X} environments (first row) have been developed by the authors to test exploration. The rest are from DeepMind Control Suite with proprioceptive observations. The proposed CMCGS shows the best performance overall.
}
\label{fig:reward_plot}
\end{figure*}

We plot the means of the episode returns in Fig.~\ref{fig:reward_summary} and the complete results in Fig.~\ref{fig:reward_plot}. The plots show the mean plus-minus two standard errors. CMCGS demonstrates the best overall performance compared to the baselines. In the four 2D environments that test exploration, CMCGS is the best, and it is also the best overall in the DMC environments where exploration is less challenging. CMCGS is highly robust to the different sizes of simulation budgets and achieves the best overall mean performance in both the 2D and DMC environments at every simulation budget. CEM uses a single bandit for each time step, MCTS-PW, VOOT, and MCTS-ABS create a search tree that grows with each time step, and CMA-ES creates a distribution over all time steps with temporal correlations but assumes a uni-modal distribution and does not divide planning into different nodes for different parts of the state distributions \citep{rubinstein1997optimization, pinneri2021sample}. Therefore, CEM and CMA-ES perform poorly in tasks that require
strong exploration, and VOOT, MCTS-PW, and MCTS-ABS perform poorly in the DMC tasks that require taking advantage of informative rewards and dealing with high-dimensional action spaces. CMCGS uses several nodes, similar to MCTS, for each layer but strongly limits the number of nodes at each time step.
CMCGS explores successfully but can also take advantage of informative rewards to find the best actions with the node policies $\pi_q(\va)$.

CMA-ES outperforms CMCGS in the \textit{cheetah} and \textit{walker} environments, most likely due to 
the covariance matrix that takes correlations over all time steps 
into account. However, the benefit of the full covariance matrix is nonexistent in other environments. CMA-ES is sometimes inferior to CEM despite CEM assuming a much simpler sampling distribution without any temporal correlations, which was also observed in \citet{duan2016benchmarking}.
The MCTS-based algorithms VOOT, MCTS-PW, and MCTS-ABS perform well in low-dimensional and sparse reward environments such as the 2d navigation and reacher environments and \textit{ball\_in\_cup-catch}, but they struggle with high-dimensional action spaces, especially in the \textit{walker} environments. We observe that the progressive widening strategy is insufficient for making the MCTS-based methods handle long planning horizons with the search space growing exponentially. For the complete results, please see Appendix~\ref{app:results}. We also compare the wall-clock efficiency of CMCGS to the baselines to show that its running times are similar to those of the other methods. The results can be found in Appendix~\ref{app:time}. 

\subsection{Large-Scale CMCGS with Learned Models}

We demonstrate the generality of CMCGS by
evaluating it with learned dynamics models in high-dimensional continuous control environments and comparing it to CEM. For evaluation, we use pixel-based environments from the DeepMind Control Suite \citep{tassa2018deepmind} and PlaNet \citep{hafner2019learning} as the underlying algorithm for learning the dynamics model. We use an open-sourced implementation of PlaNet and pre-trained models by \citet{Kaixhin2019} and replace CEM with CMCGS as the planner during evaluation. PlaNet has an encoder that learns a latent representation from pixels, and CMCGS performs planning and clustering in this learned latent space.

We copy the hyperparameters for CEM from the original paper \citep{hafner2019learning} and either perform equally many simulator interactions with CMCGS or spend equal time. We let CMCGS collect 400 trajectories in parallel to make the running time comparable to CEM. The mean episode returns in Table~\ref{tab:planet} show that CMCGS outperforms CEM in six of the seven environments for the same number of environment steps and in five of the seven environments in total and is significantly better overall. We find that CMCGS does not exploit the model inaccuracies, which makes it a robust and general planning algorithm for high-dimensional continuous control tasks.

\begin{table}
\caption{The average rewards attained in the DMC environments with image observations using learned PlaNet dynamics models plus-minus two standard errors. There are two variants of CEM. In the first variant, the number of simulator steps is equal to the number of simulator steps used by CMCGS. In the second variant of CEM, the running times of the algorithms have been matched on our workstation.}
\centering
\begin{tabular}{l S[table-format=3.1] @{${}\pm{}$} S[table-format=2.1] S[table-format=3.1] @{${}\pm{}$} S[table-format=2.1] S[table-format=3.1] @{${}\pm{}$} S[table-format=2.1]}
\toprule
Environment & \multicolumn{2}{c}{CEM (steps)} & \multicolumn{2}{c}{CEM (time)} & \multicolumn{2}{c}{CMCGS} \\
\midrule
Ball-in-Cup Catch & 713.4 & 31.6 & 728.4 & 32.8 & \textBF{844.6} & 20.6 \\
Cartpole Balance & \textBF{996.7} & 0.1 & 991.0 & 0.3 & 993.4 & 0.2 \\
Cartpole Swingup & 763.1 & 4.5 & 778.0 & 12.1 & \textBF{857.0} & 0.8 \\
Cheetah Run & 556.5 & 8.5 & 615.8 & 12.7 & \textBF{628.3} & 16.6 \\
Finger Spin & 651.1 & 9.5 & \textBF{857.7} & 2.4 & 833.9 & 3.2 \\
Reacher Easy & 828.4 & 46.1 & 832.6 & 45.7 & \textBF{887.1} & 36.1 \\
Walker Walk & 863.1 & 21.3 & 901.9 & 9.1 & \textBF{950.6} & 4.6 \\
\midrule
Mean & 767.5 & 23.2 & 815.0 & 22.6 & \textBF{856.3} & 17.1 \\
\bottomrule
\end{tabular}
\label{tab:planet}
\end{table}

\subsection{Ablations}

\begin{table}
\caption{The average rewards attained in 2D and DMC environments for simulation budget 2,500 plus-minus two standard errors for different action bandit design choices.}
\centering
\begin{tabular}{l c}
\toprule
Ablation & Result \\
\midrule
$\text{CMCGS}$ & \textBF{679.6} $\pm$ 22.5 \\

Initial action distribution uniform & 666.1 $\pm$ 21.1 \\
Uniform exploration with $p_\text{unif} = 0.05 \%$ & 671.9 $\pm$ 21.4 \\
Uniform exploration with $p_\text{unif} = 0.1 \%$ & 668.5 $\pm$ 21.8 \\
No Bayesian variance updates & 633.1 $\pm$ 26.6 \\
Reward-weighted mean & 673.3 $\pm$ 21.2 \\
VOO-sampling & 658.7 $\pm$ 23.6 \\
Greedy actions without noise, $\mathcal{E}_\text{top} = 0 $ & 670.2 $\pm$ 22.4 \\
No sampling of greedy actions, $\mathcal{N}_\text{top} = 1 $ & 647.5 $\pm$ 23.6 \\
No greedy actions, $\epsilon = 1.0$ & 649.1 $\pm$ 22.5 \\
Higher proportion of greedy actions, $\epsilon = 0.5$ & 668.7 $\pm$ 23.2 \\

\bottomrule
\end{tabular}
\label{tab:action_abls}
\end{table}

\begin{table}
\caption{The average rewards attained in 2D and DMC environments for simulation budget 2,500 plus-minus two standard errors for different clustering algorithms.}
\centering
\begin{tabular}{l c}
\toprule
Clustering algorithm & Result \\
\midrule
Hierarchical (Ward linkage) & \textBF{679.6} $\pm$ 22.5 \\
KMeans & 674.2 $\pm$ 31.2 \\
GMM & 670.8 $\pm$ 22.1 \\
\bottomrule
\end{tabular}
\label{tab:clustering_alg_abls}
\end{table}

We perform several ablation studies to analyze the impact of our design choices on the performance of CMCGS. To make the returns from our custom environments and DMC environments comparable, we scale the former by adding one and multiplying by 500. The impacts of various design choices related to the action selection are shown in Table~\ref{tab:action_abls}. Replacing the initial Gaussian distribution for $\pi_q(\va)$ with actions sampled from a uniform distribution until the first update degrades the performance. Furthermore, adding uniform exploration with probability $p_\text{unif}$ throughout the planning process hurts the performance. The Bayesian variance updates (see Eq.~\ref{eq:bayes}) are crucial for high performance and outperform CEM-like variance updates, where the new variance for $\pi_q(\va) $ is equal to the variance of the elite actions. We also evaluate discarding the CEM-inspired elite actions entirely and simply letting the means of $\pi_q(\va)$ equal the reward-weighted means of the actions in the replay buffer, but this approach fails to improve the performance of our method. We also experiment with splitting the action space into Voronoi cells \citep{kim2020monte} and sampling the greedy action from the cell corresponding to the best action found in $\mathcal{D}_q$, but it is not competitive with the current greedy action selection mechanism, where we sample one of the top actions and add noise to it.
When $\mathcal{E}_\text{top} = 0$, we do not add any noise to the greedy actions sampled from the replay buffer $\mathcal{D}_q$. When $N_\text{top} = 1$, we take the best action deterministically from $\mathcal{D}_q$.
Finally, we evaluate values of $\epsilon$ different from the chosen one, $\epsilon = 0.7$. When $\epsilon = 1$, we always sample from $\pi_q(\va)$ and never use the top actions found in $\mathcal{D}_q$. None of the ablations presented here improve on the chosen CMCGS design, but many of them are close to the original result-wise, which highlights the robustness of CMCGS. In Table~\ref{tab:clustering_alg_abls}, we compare clustering algorithms. KMeans and GMM are not significantly inferior to hierarchical clustering in terms of performance, but they are slower in terms of running time. Further ablation studies can be found in Appendix~\ref{app:ablations}.

\section{Limitations and Future Work}
In this section, we explain the main limitations of our work and how they could be addressed in future work.

\textbf{Bootstrapping}: In our current implementation, CMCGS builds the search graph from scratch at each timestep.
Model-predictive control can benefit from using the best solution from previous timestep(s) to bias the optimization process towards more promising trajectories if the model is accurate \citep{herceg2015dominant, bhardwaj2020blending}.
In CMCGS, this could be done by biasing the mean of the action policies of the search nodes towards the old best actions. A more trivial approach would be to simply inject the old best trajectory into the replay memory of the search nodes.

\textbf{Learning}: MCTS can benefit from learned value functions, policy priors, or rollout policies that have been trained using imitation learning from pre-existing datasets~\citep{silver2016mastering, kartal2019action, swiechowski2018improving}. We believe that this could improve the performance of CMCGS.

\textbf{Arbitrary graph structures}: Currently, CMCGS uses a layered directed acyclic graph to represent the state space, which does not allow the search graph to re-use previous experience in environments where the agent can navigate back and forth between different states (such as searching through a maze). This limitation could be lifted by allowing the algorithm to build arbitrary graph structures (such as complete graphs) to better utilize the structure of the state space during planning.

\textbf{Theoretical properties}: We leave the analysis of the theoretical properties of CMCGS and its convergence to future work. Note that our ablation study in Table~\ref{tab:action_abls} shows that adding uniform exploration and Voronoi optimistic sampling during the planning process has a negative impact on the performance of CMCGS in the 2D and DMC environments that we evaluated it on.

\section{Conclusion}

In this paper, we propose Continuous Monte Carlo Graph Search (CMCGS), an extension of the popular MCTS algorithm for solving decision-making problems with high-dimensional continuous state and action spaces. CMCGS builds up on the observation that different regions of the state space ask for different action bandits for estimating the reward and representing the action distribution. Based on this observation, CMCGS builds a layered search graph where, at each layer, the visited states are clustered into several stochastic action bandit nodes, which allows CMCGS to solve complex high-dimensional continuous control problems efficiently. CMCGS outperforms MCTS-PW, VOOT, MCTS-ABS, CEM, and CMA-ES
in several complex continuous control environments such as DeepMind Control Suite benchmarks. Experiments in sparse-reward custom environments indicate that CMCGS has significantly better exploration capabilities than the evaluated baselines, especially given a high-dimensional action space.

CMCGS can be efficiently parallelized, which makes the algorithm practically relevant also for large-scale planning. Our experiments with the learned PlaNet dynamics models show that CMCGS can be scaled up to be competitive with CEM as a planning component of general model-based reinforcement learning algorithms that learn from pixels, even when the number of environment interactions is not a limiting factor.
We believe that the proposed CMCGS algorithm could be a building block for a new family of Monte Carlo methods applied to decision-making problems with continuous action spaces.



\begin{acks}
We acknowledge the computational resources provided by the Aalto Science-IT project and CSC, Finnish IT Center for Science. The work was funded by Research Council of Finland (aka Academy of Finland) within the Flagship Programme, Finnish Center for Artificial Intelligence (FCAI).
J.~Pajarinen and Y.~Zhao were partly supported by Research Council of Finland (345521).
\end{acks}



\bibliographystyle{ACM-Reference-Format} 
\bibliography{rebibed}


\newpage
\clearpage
\appendix

\section{Ethics Statement}

Our work does not appear to have any immediate adverse effects on society. Our research did not involve any human subjects or participants. We only use pre-trained neural network models that have not been trained on sensitive or confidential information. Neither did we use any such information ourselves. Our model does not directly pertain to making real-world decisions involving individuals. However, our work is focused on improving continuous control, which could be applicable to, for instance, robotics. Therefore, it is, in principle, possible that our methods could be employed in unforeseen harmful ways, such as military or law enforcement robotics.

\section{Baselines}
\label{app:baselines}

We included random shooting, Cross-Entropy Method \citep[CEM]{rubinstein1997optimization}, Covariance Matrix Adaptation Evolution Strategy \citep[CMA-ES]{hansen2003reducing}, MCTS with Progressive Widening \citep[MCTS-PW]{chaslot2008progressive,couetoux2011continuous,coulom2007computing}, Voronoi Optimistic Optimization applied to trees \citep[VOOT]{kim2020monte}, and MCTS with Iteratively Refined State Abstractions \citep[MCTS-ABS]{sokota2021monte} as the baselines in our work.

Planning with random shooting relies on sampling trajectories until the complete simulation budget has been exhausted, selecting the first action of the trajectory with the highest return, and repeating this process. To make the results comparable to CMCGS, we used the same initial Gaussian distribution $\mathcal{N}(\mu = \frac{1}{2} (\va_{\min} + \va_{\max}), \sigma = \frac{1}{2} (\va_{\max} - \va_{\min}))$ for random shooting and CEM. The planning horizon is ten because that is equal to the number of steps needed to discover the reward in the \textit{2D navigation} environments.

We use the CEM \citep{rubinstein1997optimization} implementation from \citet{Pineda2021MBRL}. We sample trajectories from the same initial action distribution as CMCGS and random shooting. Then, we re-fit the action distribution $\pi(\va_t)$ to the elite samples. To make the results comparable, we use the same elite ratio as CMCGS, 0.1. The planning horizon is equal to ten, as for random shooting. For the full sample budget of 5,000 interactions, we use ten iterations with a population size of 50. For smaller budgets, we reduce the number of iterations and the population size, as we found that to outperform keeping either of the hyperparameters fixed and decreasing the other one. For the PlaNet experiments, the CEM hyperparameters were copied from the original work \citep{hafner2019learning}, except for the experiments with equal wall-clock time, where we sample more trajectories and increase the planning horizon to utilize the extra computation available.

For CMA-ES \citep{hansen1996adapting}, we used the CMA-ES python library\footnote{https://pypi.org/project/cmaes/} that has no method-specific hyperparameters to be tuned. We use CMA-ES in continuous control by creating a joint distribution over the action dimensions for each time step with a full covariance matrix. Therefore, the number of dimensions is $d \times t$, where $d$ is equal to the dimensionality of the action space and $t$ the planning horizon, which is equal to ten, the same as for CEM and random shooting. We perform ten iterations, similar to CEM. After each iteration, the joint action distribution is updated towards the best trajectories found during the previous iteration. Inspired by the performance of CMA-ES on some of the continuous control environments, we also experimented with CMA-ES with the PlaNet learned models. However, CMA-ES failed to perform adequately as it simply exploited the model inaccuracies. 

We implemented the continuous MCTS-based methods on top of the continuous MCTS implementation by \citet{hoffman2020acme}, and we performed a systematic hyperparameter search to optimize the hyperparameters of these methods. For MCTS-PW, we use the UCB \citep{kocsis2006bandit} formula for action selection:
\[
    a_t = \argmax_a Q(a) + c_\text{ucb}\sqrt{\frac{\log N}{N(a)}},
\]
where $Q(a)$ stands for the average return, when action $a$ is taken, $c_\text{ucb}$ is a hyperparameter, $N$ is the number of times the node has been visited, and $N(a)$ the number of times action $a$ has been selected in the node. The action corresponding to the highest Q-value is returned at the end of the search. To control the expansion of the search tree, we compute the maximum number of children a node can have as:
\[
    \text{num\_children}[s, a] = \lceil C_\text{pw}N^\kappa \rceil,
\]
where $C_\text{pw}$ and $\kappa$ are hyperparameters, and $N$ is the number of times the node has been visited \citep{couetoux2011continuous}. For MCTS with iteratively refined state abstractions, we use the UCB algorithm, and our implementation closely follows Algorithm~2 in \citet{sokota2021monte}, and we compute the value of $\epsilon_n$ using the formula 
\[
    \epsilon_\text{num\_visits[s, a, s'']} = A \times N^{-B},
\]
where $A$ and $B$ are hyperparameters that we tune and N is the number of times that action $a$ has been sampled in state $s$. Finally, our implementation of VOOT closely follows the description in \citet{kim2020monte}. For the purposes of efficiency, we implemented the procedure \textit{SampleBestVCell} by sampling from a Gaussian centered at the best action found so far, as the rejection sampling procedure proposed by the authors was not practically feasible due to its inefficiency in terms of running time. Then, we systematically fine-tuned the hyperparameters of VOOT. The tuned hyperparameters of the MCTS-based methods and VOOT are shown in Tables~\ref{tab:mcts_pw_hyperparams}-\ref{tab:voot_hyperparams}.

\begin{table}[h]
\caption{Hyperparameters for MCTS with progressive widening.}
\centering
\begin{tabular}{llc}
\toprule
Parameter & Explanation & Value \\
\midrule
$\gamma$ & Discount factor & 0.99 \\
$N_r$ & Rollout length & 5 \\
$c_\text{ucb}$ & Multiplier for UCB exploration term & 0.75 \\
$\kappa$ & Exponent for progressive widening & 0.6 \\
$C_\text{pw}$ & Multiplier for progressive widening & 3 \\
\bottomrule
\end{tabular}
\label{tab:mcts_pw_hyperparams}
\end{table}

\begin{table}[h]
\caption{Hyperparameters for MCTS with iteratively refined state abstractions.}
\centering
\begin{tabular}{llc}
\toprule
Parameter & Explanation & Value \\
\midrule
$\gamma$ & Discount factor & 0.99 \\
$N_r$ & Rollout length & 5 \\
$c_\text{ucb}$ & Multiplier for UCB exploration term & 0.75 \\
$A$ & Exponent for $\epsilon_\text{num\_visits}$ & 0.1 \\
$B$ & Multiplier for $\epsilon_\text{num\_visits}$ & 1 \\
\bottomrule
\end{tabular}
\label{tab:mcts_abs_hyperparams}
\end{table}

\begin{table}[h]
\caption{Hyperparameters for Voronoi optimistic optimization for trees.}
\centering
\begin{tabular}{llc}
\toprule
Parameter & Explanation & Value \\
\midrule
$\gamma$ & Discount factor & 0.99 \\
$\omega$ & Exploration probability & 0.6 \\
$H$ & Horizon & 10 \\
$N_r$ & Number of re-evaluations & 5 \\
$\kappa_r$ & Decaying factor & 0.2 \\
\bottomrule
\end{tabular}
\label{tab:voot_hyperparams}
\end{table}

\section{CMCGS Hyperparameters}
\label{app:hyperparams}

\begin{table*}[h]
\caption{CMCGS hyperparameters.}
\centering
\begin{tabular}{llccc}
\toprule
Parameter & Explanation & Toy Env. & Continuous Control & PlaNet \\
\midrule
$N_t$ & Trajectories collected in parallel & 800 & 1 & 400 \\
$|\mathcal{D}_q|_{\max}$ & Maximum replay buffer size & 1000 & 500 & 500 \\
$m$ & Depth expansion threshold & 100 & 50 & 50 \\
$\epsilon$ & Probability of using node policy $\pi_q$ & 0.5 & 0.7 & 1.0 \\
$N_{\text{top}}$& No. top actions for action selection & 50 & 3 & 20 \\
$d_{\text{init}}$ & Initial depth & 5 & 3 & 3\\
$d_{\max}$ & Maximum depth & 5 & $\infty$ & 10 \\
$N_r$ & Rollout length & 0 & 5 & 5 \\
$n_{\max}$ & Maximum number of clusters & 2 & $\infty$ & $\infty$ \\
$\alpha$ & For inverse-gamma prior & 5 & 5 & 5 \\
$\beta$ & For inverse-gamma prior & 2 & 2 & 2 \\
$r_\text{elite}$ & Proportion of elite samples for updating $\pi_q(\va)$ & 0.1 & 0.1 & 0.1 \\
$\mathcal{E}_\text{top}$ & Noise for top actions & 0.1 & 0.1 & 0.1 \\
\bottomrule
\end{tabular}
\label{tab:hyperparams}
\end{table*}

The hyperparameters used for CMCGS are shown in Table~\ref{tab:hyperparams} and included in the submitted code. For the toy environment and continuous control experiments, we have $\epsilon < 1$. When the greedy action is used (see Section~\ref{sec:method}), it is sampled uniformly from the $N_{\text{top}}$ best actions. For the PlaNet experiments, we use $\epsilon = 1$, which means that the action is always sampled from $\pi_q$. However, the search is conducted with a learned dynamics model. To prevent the exploitation of the inaccuracies in the dynamics model, we average the $N_{\text{top}}$ best actions in the replay buffer of the root node, measured in terms of the trajectory return, and the resulting action is taken.

\section{Custom 2D Environments}
\label{app:custom_envs}

\begin{figure}[tp]
\centering
\subfloat[2D navigation task using circular obstacles (\textit{2d-navigation-circles}).]
{\includegraphics[width=0.75\linewidth,trim={40mm 43mm 35mm 42mm},clip]{2d-navigation-circles.png}
\label{fig:2d-envs-circle-app}}
\\
\subfloat[2D navigation task using rectangular obstacles (\textit{2d-navigation-boxes}).]
{\includegraphics[width=0.75\linewidth,trim={40mm 43mm 35mm 42mm},clip]{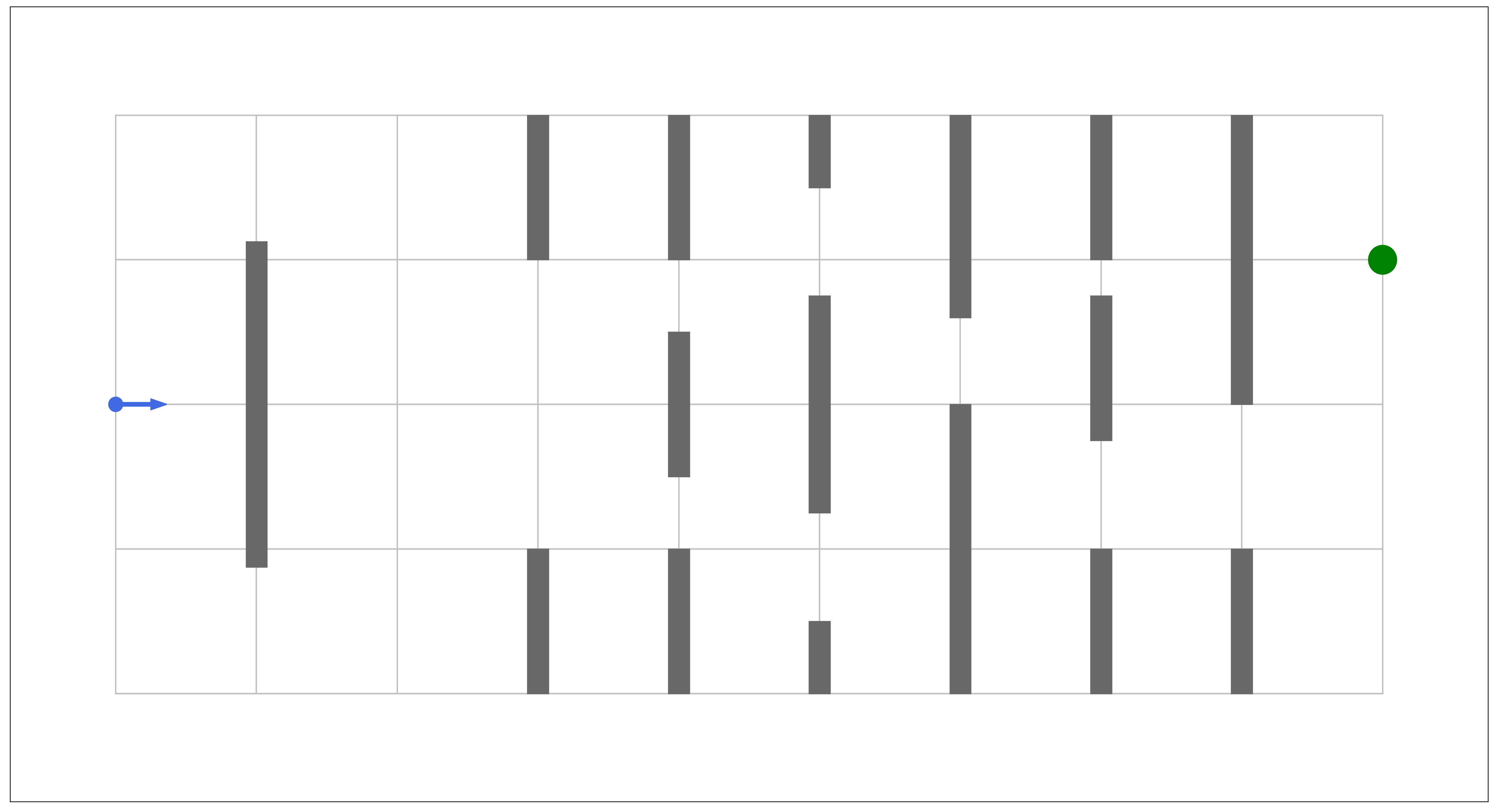}
\label{fig:2d-envs-boxes-app}}
\caption{Custom 2D navigation environments used in the experiments.}
\label{fig:2d-envs-app}
\end{figure}

We evaluate CMCGS and the chosen baselines on four 2D environments designed by us. These environments are exploration-heavy, and chosen to counterbalance the impact of the environments from DeepMind Control Suite with mostly informative rewards \citep{tassa2018deepmind}.

The 2D navigation environments with circles and rectangular are shown in Figure~\ref{fig:2d-envs-app}. The environments are otherwise similar, and the only difference is the shape of the obstacles. The agent applies a new action when it crosses a vertical line. The episode terminates if the agent hits an obstacle. The action space is $[-1, 1]$, and the action represents a force applied to the object in the vertical direction. The horizontal component of the velocity is assumed to be constant. The state space is $[-1, 1]^3$ and consists of the x-coordinate, y-coordinate, and the velocity of the object. The agent receives a big reward for reaching the goal, the green dot, and a big penalty if it hits an obstacle. Furthermore, the agent receives a small bonus for every timestep it has not hit an obstacle and a small penalty that depends on the magnitude of the action. The penalty encourages actions closer to zero.

\begin{figure}[tp]
\centering
\includegraphics[width=0.75\linewidth,trim={35mm 15mm 30mm 20mm},clip]{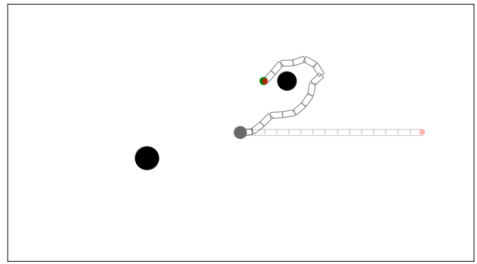}
\caption{2D reacher task using $15$-linked arm (\textit{2d-reacher-fifteen-poles})}
\label{fig:2d-reacher-fifteen-app}
\end{figure}

The 2D Reacher environment with a 15-linked arm is shown in Figure~\ref{fig:2d-reacher-fifteen-app}. The 2D Reacher with a 30-linked arm only differs in the number of links. The action space of the environments is $[-1, 1]^d$, where $d$ is the number of links. The action represents the forces applied to the joints. The state with dimension $2 \cdot d$ consists of the current angles and velocities of the joints. The reward is sparse. The agent receives a reward of one if it reaches the goal and zero otherwise. The episode terminates when the goal has been reached or the maximum number of actions, 200, has been taken. Hitting an obstacle does not terminate the episode.

\section{Ablation Studies}
\label{app:ablations}

\begin{table}
\caption{The average rewards attained in the 2D and DMC environments with search budget 2,500 plus-minus two standard errors for different values of $m$, the CMCGS depth expansion threshold.}
\centering
\begin{tabular}{c c}
\toprule
Depth expansion threshold & Result \\
\midrule
25 & 664.0 $\pm$ 21.2 \\
33 & 665.1 $\pm$ 22.0 \\
50 & \textBF{679.6} $\pm$ 22.5 \\
100 & 663.2 $\pm$ 22.9 \\
\bottomrule
\end{tabular}
\label{tab:cmcgs_threshold}
\end{table}

\begin{table}
\caption{The average rewards attained in the 2D and DMC environments with search budget 2,500 plus-minus two standard errors for different values of $N_r$, the CMCGS rollout lengths.}
\centering
\begin{tabular}{c c}
\toprule
Rollout length & Result \\
\midrule
0 & 609.7 $\pm$ 25.1 \\
1 & 621.0 $\pm$ 24.7 \\
3 & 659.2 $\pm$ 24.3 \\
5 & \textBF{679.6} $\pm$ 22.5 \\
10 & 668.9 $\pm$ 20.4 \\
\bottomrule
\end{tabular}
\label{tab:cmcgs_rollout_len}
\end{table}

\begin{table}
\caption{The average rewards attained in the 2D and DMC environments with search budget 2,500 plus-minus two standard errors for different values of $d_\text{init}$, the CMCGS initial graph depth.}
\centering
\begin{tabular}{c c}
\toprule
Initial graph depth & Result \\
\midrule
1 & 654.6 $\pm$ 23.8 \\
2 & 660.8 $\pm$ 22.9 \\
3 & \textBF{679.6} $\pm$ 22.5 \\
5 & 671.8 $\pm$ 20.7 \\
8 & 669.7 $\pm$ 21.1 \\
10 & 661.9 $\pm$ 21.5 \\
\bottomrule
\end{tabular}
\label{tab:cmcgs_initial_graph_depth}
\end{table}

\begin{table}
\caption{The average rewards attained in the 2D and DMC environments with search budget 2,500 plus-minus two standard errors for different covariance matrices of the state distributions $p_q(\vs)$.}
\centering
\begin{tabular}{c c}
\toprule
Type of covariance matrix & Result \\
\midrule
Full covariance & 671.1 $\pm$ 24.1 \\
Diagonal covariance & \textBF{679.6} $\pm$ 22.5 \\
\bottomrule
\end{tabular}
\label{tab:cmcgs_covar_matrix}
\end{table}

\begin{table}
\caption{The average rewards attained in PlaNet environments plus-minus two standard errors for different linkage metrics for hierarchical clustering.}
\centering
\begin{tabular}{l c}
\toprule
Linkage & Result \\
\midrule
Ward & \textBF{852.9} $\pm$ 19.5 \\
Average & \textBF{852.9} $\pm$ 19.0 \\
Complete & 851.0 $\pm$ 19.0 \\
Single & 718.3 $\pm$ 46.9 \\
\bottomrule
\end{tabular}
\label{tab:clustering_linkage_abls}
\end{table}

The depth expansion parameter $m$ of CMCGS determines how many samples the final layer of the CMCGS must have collected before a new layer is added to the graph. The impact of the value of $m$ on the performance is shown in Table~\ref{tab:cmcgs_threshold}. We found that the algorithm is not particularly sensitive to the choice of $m$, but the chosen value of $50$ performs the best. The results in Table~\ref{tab:cmcgs_rollout_len} show that performing rollouts of a suitable length has a significant positive impact on the performance of CMCGS in the continuous control setting, and the results in Table~\ref{tab:cmcgs_initial_graph_depth} indicate that initializing the graph with more layers than one can help improve the performance in the same setting. Finally, we analyzed the impact of the covariance matrix of the node state distribution $p_q(\vs_t)$. Our results suggest that using a full covariance matrix fails to have a positive overall impact on the results, so we used a diagonal covariance matrix for $p_q(\vs_t)$ in all of our experiments. 

We show the impact of the hierarchical clustering linkage function on the performance in the PlaNet setting in Table~\ref{tab:clustering_linkage_abls}. We evaluate the performance with 120,000 environment interactions and the pre-trained dynamics models. The results show that the choice between Ward, average, and complete linkage is unimportant in this setting, but the single linkage should not be chosen. Note that the result for Ward linkage is slightly different from that in Table~\ref{tab:planet} because we used an initial, untuned value of $\beta = 1$ in this experiment, and we could not re-run the experiment using the final, tuned value of $\beta = 2$ due to limited access to computational resources.

\section{Code, Assets, and Reproducibility}

\begin{table*}[ht]
\caption{The average running times per iteration attained in \textit{2D Reacher}, \textit{2D Navigation}, and \textit{walker-walk} from DeepMind Control Suite plus-minus two standard errors.}
\centering
\begin{tabular}{l c c c c}
\toprule
Algorithm & \textit{2d-navigation-circles} & \textit{2d-reacher-thirty-poles} & \textit{walker-walk} & Average (s) \\
\midrule
Random shooting & \textBF{2.94} $\pm$ 0.05 & 1.59 $\pm$ 0.06 & 1.77 $\pm$ 0.06 & \textBF{2.10} $\pm$ 0.06 \\
VOOT \citep{kim2020monte} & 3.11  $\pm$ 0.03 & 1.68 $\pm$ 0.07 & \textBF{1.71} $\pm$ 0.05 & 2.17 $\pm$ 0.06 \\
MCTS-PW \citep{chaslot2008progressive} & 3.39 $\pm$ 0.04 & \textBF{1.57} $\pm$ 0.04 & 2.18 $\pm$ 0.04 & 2.38 $\pm$ 0.04 \\
CEM \citep{rubinstein1997optimization} & 3.45 $\pm$ 0.02 & 2.10 $\pm$ 0.17 & 1.79 $\pm$ 0.06 & 2.44 $\pm$ 0.11 \\
CMCGS (ours) & 3.50 $\pm$ 0.02 & 2.57 $\pm$ 0.14 & 1.87 $\pm$ 0.05 & 2.64 $\pm$ 0.08 \\
CMA-ES \citep{hansen1996adapting} & 3.25 $\pm$ 0.05 & 2.40 $\pm$ 0.09 & 4.44 $\pm$ 0.16 & 3.36 $\pm$ 0.11 \\
MCTS-ABS \citep{sokota2021monte} & 3.52 $\pm$ 0.07 & 1.86 $\pm$ 0.08 & 6.65 $\pm$ 0.07 & 4.01 $\pm$ 0.07 \\
\bottomrule
\end{tabular}
\label{tab:running_time}
\end{table*}

We have open-sourced our implementation of CMCGS on GitHub\footnote{https://github.com/kallekku/cmcgs}.
We used the PlaNet \citep{hafner2019learning} implementation of \citet{Kaixhin2019} and their pre-trained model weights (MIT license). Our implementation of CMCGS is also built on Google DeepMind Acme \citep{hoffman2020acme}.

\section{Infrastructure and Compute}

The toy environment experiments were run on a workstation with a 24-core CPU and 32 GB of memory and took less than 10 minutes. No GPU was used for these experiments. The continuous control and PlaNet experiments were run on an HPC cluster. For evaluating CMCGS with limited environment interaction, we performed 7 * 10 * 12 = 840 runs (7 methods, 10 budgets, and 12 environments). Each of these runs used 10 CPU cores with 3 GB of RAM per core. No GPUs were used for these runs. Most runs lasted between one and 24 hours.
For the PlaNet experiments, we performed 3 * 20 = 60 runs (3 methods, 20 seeds, and each run evaluated the method on all environments). All runs lasted fewer than 24 hours, and they used one V100 GPU with 32 GB of GDDR SDRAM and 6 CPU workers per GPU. In addition to these runs, we performed numerous ablation studies and initial experiments during the development of our method.

\section{Running Times}
\label{app:time}

We compared the running times of CMCGS to the baseline methods in three different environments. The average running times per iteration with a search budget of 2,500 are shown in Table~\ref{tab:running_time}. The results show that the running time of CMCGS is highly competitive with those of the baselines and clearly superior to that of MCTS-ABS and CMA-ES. The clustering in CMCGS is not highly time-consuming because of the limitations on how often clustering is attempted and the relatively small size of the replay memory of CMCGS.

\newgeometry{left=2cm,right=2cm,top=3cm,bottom=3cm}

\begin{landscape}

\section{Full Results on Continuous Control with Limited Interaction}
\label{app:results}

\subsection{CMCGS}
\begin{table}[h]
\caption{Full results for CMCGS on continuous control with limited interaction: mean plus-minus two standard errors.}
\centering
\begin{tabular}{l S[table-format=3.2] @{${}\pm{}$} S[table-format=2.2] S[table-format=3.2] @{${}\pm{}$} S[table-format=2.2] S[table-format=3.2] @{${}\pm{}$} S[table-format=2.2] S[table-format=3.2] @{${}\pm{}$} S[table-format=2.2] S[table-format=3.2] @{${}\pm{}$} S[table-format=2.2] S[table-format=3.2] @{${}\pm{}$} S[table-format=2.2] } 
\toprule
Environment & \multicolumn{2}{c}{0,5K} & \multicolumn{2}{c}{1K} & \multicolumn{2}{c}{1,5K} & \multicolumn{2}{c}{2K} & \multicolumn{2}{c}{2,5K} & \multicolumn{2}{c}{3K} \\
\midrule
2d-navigation-circles & 0.95 & 0.01 & 0.97 & 0.00 & 0.97 & 0.00 & 0.97 & 0.00 & 0.97 & 0.00 & 0.97 & 0.00 \\
2d-navigation-boxes & 0.59 & 0.02 & 0.66 & 0.02 & 0.70 & 0.02 & 0.73 & 0.02 & 0.74 & 0.02 & 0.76 & 0.02 \\
2d-reacher-fifteen-poles & 0.31 & 0.04 & 0.39 & 0.04 & 0.37 & 0.04 & 0.42 & 0.04 & 0.46 & 0.04 & 0.44 & 0.04 \\
2d-reacher-thirty-poles & 0.34 & 0.04 & 0.42 & 0.04 & 0.44 & 0.04 & 0.48 & 0.04 & 0.53 & 0.04 & 0.54 & 0.04 \\
ball\_in\_cup-catch & 922.07 & 6.17 & 943.04 & 4.49 & 957.62 & 3.04 & 965.16 & 2.56 & 968.42 & 2.35 & 971.99 & 2.14 \\
cartpole-balance\_sparse & 419.37 & 24.79 & 569.36 & 27.10 & 704.80 & 28.76 & 780.76 & 27.26 & 855.57 & 23.10 & 873.88 & 22.87 \\
cartpole-swingup & 543.98 & 7.04 & 640.43 & 6.62 & 700.39 & 5.75 & 728.21 & 4.95 & 744.50 & 5.14 & 760.67 & 4.48 \\
finger-spin & 185.44 & 4.35 & 222.21 & 4.81 & 246.01 & 5.32 & 259.57 & 5.58 & 281.02 & 6.13 & 281.62 & 5.82 \\
reacher-hard & 720.26 & 12.83 & 775.18 & 9.40 & 816.72 & 8.00 & 845.79 & 6.81 & 865.80 & 6.02 & 881.67 & 5.59 \\
cheetah-run & 369.15 & 5.11 & 422.69 & 4.14 & 481.22 & 4.20 & 519.20 & 4.50 & 547.24 & 4.79 & 568.19 & 4.93 \\
walker-walk & 283.62 & 4.82 & 298.53 & 4.38 & 308.67 & 4.39 & 309.68 & 4.53 & 309.91 & 3.97 & 315.47 & 4.56 \\
walker-run & 112.46 & 3.93 & 130.78 & 3.92 & 141.84 & 3.94 & 155.05 & 3.78 & 157.77 & 4.34 & 164.09 & 3.98 \\
\bottomrule
\end{tabular}
\label{mcgs-full-results-a}
\end{table}

\begin{table}[h]
\caption{Full results for CMCGS on continuous control with limited interaction: mean plus-minus two standard errors.}
\centering
\begin{tabular}{l S[table-format=3.2] @{${}\pm{}$} S[table-format=2.2] S[table-format=3.2] @{${}\pm{}$} S[table-format=2.2] S[table-format=3.2] @{${}\pm{}$} S[table-format=2.2] S[table-format=3.2] @{${}\pm{}$} S[table-format=2.2] } 
\toprule
Environment & \multicolumn{2}{c}{3,5K} & \multicolumn{2}{c}{4K} & \multicolumn{2}{c}{4,5K} & \multicolumn{2}{c}{5K} \\
\midrule
2d-navigation-circles & 0.97 & 0.00 & 0.97 & 0.00 & 0.98 & 0.00 & 0.98 & 0.00 \\
2d-navigation-boxes & 0.78 & 0.02 & 0.79 & 0.02 & 0.80 & 0.02 & 0.82 & 0.02 \\
2d-reacher-fifteen-poles & 0.41 & 0.04 & 0.45 & 0.04 & 0.42 & 0.04 & 0.46 & 0.04 \\
2d-reacher-thirty-poles & 0.55 & 0.04 & 0.53 & 0.04 & 0.54 & 0.04 & 0.55 & 0.04 \\
ball\_in\_cup-catch & 972.84 & 2.17 & 974.70 & 2.02 & 975.93 & 1.85 & 975.71 & 1.92 \\
cartpole-balance\_sparse & 919.74 & 17.88 & 931.78 & 16.76 & 941.99 & 15.65 & 951.15 & 14.47 \\
cartpole-swingup & 768.46 & 3.72 & 775.41 & 4.07 & 781.45 & 3.50 & 788.05 & 3.46 \\
finger-spin & 287.49 & 6.95 & 293.12 & 6.17 & 299.73 & 6.39 & 296.94 & 6.47 \\
reacher-hard & 890.76 & 4.88 & 896.57 & 5.35 & 903.41 & 4.82 & 911.32 & 3.99 \\
cheetah-run & 584.71 & 4.74 & 604.62 & 4.80 & 618.76 & 4.86 & 625.57 & 4.75 \\
walker-walk & 320.51 & 4.86 & 312.82 & 4.69 & 325.07 & 4.54 & 322.63 & 4.96 \\
walker-run & 168.62 & 4.06 & 171.89 & 3.96 & 173.98 & 4.23 & 179.22 & 4.01 \\
\bottomrule
\end{tabular}
\label{mcgs-full-results-b}
\end{table}

\pagebreak

\subsection{CEM}
\begin{table}[h]
\caption{Full results for CEM on continuous control with limited interaction: mean plus-minus two standard errors.}
\centering
\begin{tabular}{l S[table-format=3.2] @{${}\pm{}$} S[table-format=2.2] S[table-format=3.2] @{${}\pm{}$} S[table-format=2.2] S[table-format=3.2] @{${}\pm{}$} S[table-format=2.2] S[table-format=3.2] @{${}\pm{}$} S[table-format=2.2] S[table-format=3.2] @{${}\pm{}$} S[table-format=2.2] S[table-format=3.2] @{${}\pm{}$} S[table-format=2.2] } 
\toprule
Environment & \multicolumn{2}{c}{0,5K} & \multicolumn{2}{c}{1K} & \multicolumn{2}{c}{1,5K} & \multicolumn{2}{c}{2K} & \multicolumn{2}{c}{2,5K} & \multicolumn{2}{c}{3K} \\
\midrule
2d-navigation-circles & 0.39 & 0.03 & 0.53 & 0.02 & 0.57 & 0.01 & 0.57 & 0.01 & 0.58 & 0.01 & 0.58 & 0.01 \\
2d-navigation-boxes & 0.13 & 0.02 & 0.12 & 0.02 & 0.11 & 0.01 & 0.13 & 0.02 & 0.12 & 0.02 & 0.11 & 0.01 \\
2d-reacher-fifteen-poles & 0.03 & 0.02 & 0.07 & 0.04 & 0.04 & 0.03 & 0.05 & 0.03 & 0.05 & 0.03 & 0.05 & 0.03 \\
2d-reacher-thirty-poles & 0.06 & 0.03 & 0.07 & 0.04 & 0.05 & 0.03 & 0.06 & 0.03 & 0.08 & 0.04 & 0.09 & 0.04 \\
ball\_in\_cup-catch & 330.42 & 90.18 & 721.33 & 67.67 & 836.54 & 46.03 & 865.24 & 41.24 & 899.62 & 30.61 & 917.54 & 22.26 \\
cartpole-balance\_sparse & 575.12 & 48.64 & 596.90 & 51.68 & 704.00 & 51.20 & 758.40 & 50.31 & 799.84 & 46.29 & 827.84 & 42.92 \\
cartpole-swingup & 620.23 & 12.94 & 655.57 & 10.27 & 679.89 & 11.00 & 676.15 & 10.33 & 689.57 & 13.04 & 694.91 & 13.96 \\
finger-spin & 22.32 & 2.23 & 114.70 & 6.78 & 179.08 & 8.76 & 207.68 & 10.71 & 219.63 & 9.42 & 230.14 & 9.75 \\
reacher-hard & 719.92 & 59.87 & 774.13 & 55.29 & 805.46 & 51.66 & 830.10 & 42.21 & 853.45 & 35.66 & 816.69 & 48.26 \\
cheetah-run & 105.39 & 3.81 & 259.85 & 4.54 & 321.67 & 6.00 & 376.79 & 7.42 & 424.01 & 9.21 & 459.88 & 9.26 \\
walker-walk & 95.52 & 11.00 & 255.65 & 15.89 & 292.61 & 12.12 & 321.05 & 8.00 & 311.47 & 7.70 & 316.53 & 8.87 \\
walker-run & 54.21 & 3.57 & 89.65 & 6.23 & 111.59 & 7.50 & 126.72 & 8.05 & 134.39 & 7.69 & 142.41 & 7.91 \\
\bottomrule
\end{tabular}
\label{cem-full-results-a}
\end{table}

\begin{table}[h]
\caption{Full results for CEM on continuous control with limited interaction: mean plus-minus two standard errors.}
\centering
\begin{tabular}{l S[table-format=3.2] @{${}\pm{}$} S[table-format=2.2] S[table-format=3.2] @{${}\pm{}$} S[table-format=2.2] S[table-format=3.2] @{${}\pm{}$} S[table-format=2.2] S[table-format=3.2] @{${}\pm{}$} S[table-format=2.2] } 
\toprule
Environment & \multicolumn{2}{c}{3,5K} & \multicolumn{2}{c}{4K} & \multicolumn{2}{c}{4,5K} & \multicolumn{2}{c}{5K} \\
\midrule
2d-navigation-circles & 0.59 & 0.01 & 0.58 & 0.01 & 0.57 & 0.01 & 0.58 & 0.01 \\
2d-navigation-boxes & 0.12 & 0.02 & 0.11 & 0.02 & 0.14 & 0.02 & 0.11 & 0.01 \\
2d-reacher-fifteen-poles & 0.07 & 0.04 & 0.06 & 0.03 & 0.07 & 0.04 & 0.14 & 0.05 \\
2d-reacher-thirty-poles & 0.05 & 0.03 & 0.08 & 0.04 & 0.08 & 0.04 & 0.05 & 0.03 \\
ball\_in\_cup-catch & 918.38 & 21.78 & 926.95 & 19.82 & 910.18 & 26.72 & 941.16 & 16.09 \\
cartpole-balance\_sparse & 856.83 & 41.43 & 854.63 & 41.67 & 837.68 & 44.26 & 851.87 & 39.71 \\
cartpole-swingup & 699.89 & 15.00 & 720.85 & 13.50 & 729.82 & 15.09 & 722.87 & 15.22 \\
finger-spin & 237.99 & 10.48 & 239.23 & 9.39 & 244.84 & 10.33 & 242.47 & 7.96 \\
reacher-hard & 866.78 & 35.59 & 856.65 & 34.60 & 887.47 & 26.85 & 865.71 & 35.12 \\
cheetah-run & 492.37 & 7.91 & 525.06 & 7.90 & 528.55 & 8.44 & 543.15 & 6.91 \\
walker-walk & 313.44 & 8.63 & 314.85 & 7.32 & 314.47 & 7.74 & 312.10 & 7.37 \\
walker-run & 142.29 & 7.93 & 147.09 & 6.90 & 150.20 & 7.76 & 151.51 & 7.73 \\
\bottomrule
\end{tabular}
\label{cem-full-results-b}
\end{table}

\pagebreak

\subsection{MCTS-PW}
\begin{table}[h]
\caption{Full results for MCTS-PW on continuous control with limited interaction: mean plus-minus two standard errors.}
\centering
\begin{tabular}{l S[table-format=3.2] @{${}\pm{}$} S[table-format=2.2] S[table-format=3.2] @{${}\pm{}$} S[table-format=2.2] S[table-format=3.2] @{${}\pm{}$} S[table-format=2.2] S[table-format=3.2] @{${}\pm{}$} S[table-format=2.2] S[table-format=3.2] @{${}\pm{}$} S[table-format=2.2] S[table-format=3.2] @{${}\pm{}$} S[table-format=2.2] } 
\toprule
Environment & \multicolumn{2}{c}{0,5K} & \multicolumn{2}{c}{1K} & \multicolumn{2}{c}{1,5K} & \multicolumn{2}{c}{2K} & \multicolumn{2}{c}{2,5K} & \multicolumn{2}{c}{3K} \\
\midrule
2d-navigation-circles & 0.97 & 0.00 & 0.97 & 0.00 & 0.97 & 0.00 & 0.97 & 0.00 & 0.97 & 0.00 & 0.97 & 0.00 \\
2d-navigation-boxes & 0.56 & 0.04 & 0.63 & 0.04 & 0.65 & 0.04 & 0.61 & 0.04 & 0.64 & 0.04 & 0.64 & 0.04 \\
2d-reacher-fifteen-poles & 0.12 & 0.05 & 0.14 & 0.05 & 0.17 & 0.05 & 0.19 & 0.06 & 0.16 & 0.05 & 0.14 & 0.05 \\
2d-reacher-thirty-poles & 0.21 & 0.06 & 0.19 & 0.06 & 0.22 & 0.06 & 0.20 & 0.06 & 0.25 & 0.06 & 0.22 & 0.06 \\
ball\_in\_cup-catch & 813.36 & 45.33 & 870.27 & 39.46 & 918.39 & 20.37 & 912.04 & 26.73 & 942.32 & 13.54 & 940.01 & 14.62 \\
cartpole-balance\_sparse & 353.07 & 30.23 & 468.53 & 44.20 & 451.99 & 40.58 & 473.34 & 42.77 & 479.24 & 45.44 & 478.52 & 44.43 \\
cartpole-swingup & 469.74 & 13.79 & 485.18 & 12.62 & 472.76 & 13.72 & 470.38 & 14.82 & 486.34 & 11.54 & 481.15 & 12.92 \\
finger-spin & 50.91 & 4.76 & 77.20 & 5.84 & 98.08 & 7.03 & 116.98 & 7.18 & 138.03 & 7.45 & 141.03 & 7.21 \\
reacher-hard & 687.29 & 65.78 & 794.33 & 48.20 & 755.23 & 57.26 & 798.49 & 50.43 & 757.07 & 57.03 & 797.61 & 53.97 \\
cheetah-run & 247.96 & 3.10 & 271.51 & 4.56 & 278.84 & 4.59 & 285.19 & 5.20 & 292.39 & 5.83 & 297.30 & 4.81 \\
walker-walk & 111.86 & 13.64 & 123.63 & 15.07 & 146.20 & 15.72 & 138.84 & 15.94 & 143.29 & 17.01 & 153.53 & 17.04 \\
walker-run & 49.21 & 4.85 & 49.88 & 4.88 & 57.76 & 6.36 & 57.72 & 6.44 & 55.48 & 5.72 & 57.56 & 6.01 \\
\bottomrule
\end{tabular}
\label{mcts_pw-full-results-a}
\end{table}

\begin{table}[h]
\caption{Full results for MCTS-PW on continuous control with limited interaction: mean plus-minus two standard errors.}
\centering
\begin{tabular}{l S[table-format=3.2] @{${}\pm{}$} S[table-format=2.2] S[table-format=3.2] @{${}\pm{}$} S[table-format=2.2] S[table-format=3.2] @{${}\pm{}$} S[table-format=2.2] S[table-format=3.2] @{${}\pm{}$} S[table-format=2.2] } 
\toprule
Environment & \multicolumn{2}{c}{3,5K} & \multicolumn{2}{c}{4K} & \multicolumn{2}{c}{4,5K} & \multicolumn{2}{c}{5K} \\
\midrule
2d-navigation-circles & 0.97 & 0.00 & 0.97 & 0.00 & 0.97 & 0.00 & 0.97 & 0.00 \\
2d-navigation-boxes & 0.64 & 0.04 & 0.66 & 0.04 & 0.67 & 0.04 & 0.65 & 0.04 \\
2d-reacher-fifteen-poles & 0.19 & 0.05 & 0.20 & 0.06 & 0.19 & 0.05 & 0.18 & 0.05 \\
2d-reacher-thirty-poles & 0.19 & 0.05 & 0.20 & 0.06 & 0.20 & 0.06 & 0.21 & 0.06 \\
ball\_in\_cup-catch & 943.15 & 14.35 & 949.77 & 11.14 & 940.36 & 17.60 & 955.53 & 11.33 \\
cartpole-balance\_sparse & 490.52 & 45.59 & 490.34 & 43.38 & 473.19 & 44.00 & 499.11 & 43.79 \\
cartpole-swingup & 477.37 & 12.14 & 483.42 & 13.36 & 474.28 & 10.89 & 490.76 & 14.17 \\
finger-spin & 156.77 & 7.54 & 162.16 & 6.53 & 165.10 & 7.34 & 173.07 & 6.86 \\
reacher-hard & 831.71 & 40.24 & 816.06 & 49.72 & 801.44 & 49.19 & 794.57 & 54.06 \\
cheetah-run & 294.87 & 5.19 & 299.78 & 5.33 & 299.62 & 5.75 & 301.79 & 4.89 \\
walker-walk & 145.40 & 15.76 & 137.81 & 15.56 & 143.78 & 15.46 & 170.24 & 16.88 \\
walker-run & 55.84 & 6.02 & 57.88 & 6.35 & 55.06 & 5.85 & 55.45 & 5.82 \\
\bottomrule
\end{tabular}
\label{mcts_pw-full-results-b}
\end{table}

\pagebreak

\subsection{Random Shooting}
\begin{table}[h]
\caption{Full results for Random Shooting on continuous control with limited interaction: mean plus-minus two standard errors.}
\centering
\begin{tabular}{l S[table-format=3.2] @{${}\pm{}$} S[table-format=2.2] S[table-format=3.2] @{${}\pm{}$} S[table-format=2.2] S[table-format=3.2] @{${}\pm{}$} S[table-format=2.2] S[table-format=3.2] @{${}\pm{}$} S[table-format=2.2] S[table-format=3.2] @{${}\pm{}$} S[table-format=2.2] S[table-format=3.2] @{${}\pm{}$} S[table-format=2.2] } 
\toprule
Environment & \multicolumn{2}{c}{0,5K} & \multicolumn{2}{c}{1K} & \multicolumn{2}{c}{1,5K} & \multicolumn{2}{c}{2K} & \multicolumn{2}{c}{2,5K} & \multicolumn{2}{c}{3K} \\
\midrule
2d-navigation-circles & 0.54 & 0.02 & 0.53 & 0.02 & 0.53 & 0.02 & 0.56 & 0.02 & 0.55 & 0.02 & 0.55 & 0.02 \\
2d-navigation-boxes & 0.33 & 0.02 & 0.35 & 0.02 & 0.34 & 0.02 & 0.32 & 0.02 & 0.31 & 0.02 & 0.33 & 0.02 \\
2d-reacher-fifteen-poles & 0.12 & 0.05 & 0.09 & 0.04 & 0.08 & 0.04 & 0.14 & 0.05 & 0.05 & 0.03 & 0.14 & 0.05 \\
2d-reacher-thirty-poles & 0.09 & 0.04 & 0.09 & 0.04 & 0.07 & 0.04 & 0.09 & 0.04 & 0.06 & 0.03 & 0.09 & 0.04 \\
ball\_in\_cup-catch & 937.03 & 10.02 & 959.79 & 5.37 & 963.08 & 5.89 & 968.12 & 4.21 & 970.44 & 4.07 & 971.61 & 3.89 \\
cartpole-balance\_sparse & 464.96 & 39.98 & 646.04 & 45.89 & 760.54 & 43.34 & 856.40 & 37.06 & 857.08 & 38.90 & 902.18 & 28.60 \\
cartpole-swingup & 595.44 & 11.95 & 613.71 & 11.65 & 629.75 & 12.54 & 645.31 & 12.55 & 645.52 & 12.56 & 648.04 & 11.75 \\
finger-spin & 160.98 & 7.12 & 197.48 & 8.33 & 217.34 & 10.78 & 232.13 & 10.20 & 238.97 & 8.95 & 232.79 & 10.20 \\
reacher-hard & 572.72 & 19.17 & 645.48 & 17.13 & 671.58 & 16.26 & 692.90 & 16.99 & 711.46 & 15.86 & 725.46 & 14.11 \\
cheetah-run & 290.64 & 6.40 & 320.53 & 5.96 & 341.17 & 5.98 & 352.57 & 5.62 & 360.96 & 7.30 & 367.19 & 6.70 \\
walker-walk & 267.95 & 6.86 & 269.58 & 6.00 & 279.20 & 7.40 & 280.79 & 7.09 & 280.45 & 7.95 & 279.15 & 6.26 \\
walker-run & 104.97 & 4.24 & 112.05 & 4.57 & 122.44 & 4.67 & 124.62 & 3.80 & 128.99 & 3.81 & 129.11 & 3.97 \\
\bottomrule
\end{tabular}
\label{rs-full-results-a}
\end{table}

\begin{table}[h]
\caption{Full results for Random Shooting on continuous control with limited interaction: mean plus-minus two standard errors.}
\centering
\begin{tabular}{l S[table-format=3.2] @{${}\pm{}$} S[table-format=2.2] S[table-format=3.2] @{${}\pm{}$} S[table-format=2.2] S[table-format=3.2] @{${}\pm{}$} S[table-format=2.2] S[table-format=3.2] @{${}\pm{}$} S[table-format=2.2] } 
\toprule
Environment & \multicolumn{2}{c}{3,5K} & \multicolumn{2}{c}{4K} & \multicolumn{2}{c}{4,5K} & \multicolumn{2}{c}{5K} \\
\midrule
2d-navigation-circles & 0.57 & 0.02 & 0.56 & 0.02 & 0.56 & 0.01 & 0.57 & 0.01 \\
2d-navigation-boxes & 0.32 & 0.02 & 0.31 & 0.02 & 0.30 & 0.02 & 0.29 & 0.02 \\
2d-reacher-fifteen-poles & 0.11 & 0.04 & 0.12 & 0.05 & 0.07 & 0.04 & 0.08 & 0.04 \\
2d-reacher-thirty-poles & 0.08 & 0.04 & 0.12 & 0.05 & 0.13 & 0.05 & 0.08 & 0.04 \\
ball\_in\_cup-catch & 973.59 & 3.49 & 974.41 & 3.40 & 973.34 & 3.86 & 974.87 & 3.36 \\
cartpole-balance\_sparse & 915.49 & 27.87 & 906.85 & 30.93 & 951.69 & 19.80 & 959.39 & 19.82 \\
cartpole-swingup & 648.48 & 12.85 & 660.01 & 11.86 & 653.24 & 12.83 & 650.19 & 11.99 \\
finger-spin & 230.52 & 9.85 & 250.40 & 10.43 & 241.56 & 9.47 & 245.88 & 10.29 \\
reacher-hard & 725.64 & 15.54 & 746.11 & 13.99 & 757.15 & 13.73 & 756.32 & 14.23 \\
cheetah-run & 371.18 & 6.65 & 377.33 & 8.37 & 387.32 & 6.74 & 383.75 & 6.82 \\
walker-walk & 278.65 & 6.58 & 276.61 & 7.05 & 272.28 & 7.49 & 276.11 & 7.41 \\
walker-run & 131.93 & 4.46 & 132.66 & 4.40 & 137.39 & 3.06 & 140.43 & 3.87 \\
\bottomrule
\end{tabular}
\label{rs-full-results-b}
\end{table}

\pagebreak

\subsection{CMA-ES}
\begin{table}[h]
\caption{Full results for CMA-ES on continuous control with limited interaction: mean plus-minus two standard errors.}
\centering
\begin{tabular}{l S[table-format=3.2] @{${}\pm{}$} S[table-format=2.2] S[table-format=3.2] @{${}\pm{}$} S[table-format=2.2] S[table-format=3.2] @{${}\pm{}$} S[table-format=2.2] S[table-format=3.2] @{${}\pm{}$} S[table-format=2.2] S[table-format=3.2] @{${}\pm{}$} S[table-format=2.2] S[table-format=3.2] @{${}\pm{}$} S[table-format=2.2] } 
\toprule
Environment & \multicolumn{2}{c}{0,5K} & \multicolumn{2}{c}{1K} & \multicolumn{2}{c}{1,5K} & \multicolumn{2}{c}{2K} & \multicolumn{2}{c}{2,5K} & \multicolumn{2}{c}{3K} \\
\midrule
2d-navigation-circles & 0.23 & 0.03 & 0.36 & 0.03 & 0.40 & 0.03 & 0.42 & 0.03 & 0.45 & 0.02 & 0.47 & 0.02 \\
2d-navigation-boxes & 0.09 & 0.02 & 0.21 & 0.03 & 0.17 & 0.02 & 0.19 & 0.03 & 0.20 & 0.03 & 0.21 & 0.03 \\
2d-reacher-fifteen-poles & 0.08 & 0.04 & 0.09 & 0.04 & 0.08 & 0.04 & 0.08 & 0.04 & 0.12 & 0.05 & 0.09 & 0.04 \\
2d-reacher-thirty-poles & 0.10 & 0.04 & 0.09 & 0.04 & 0.08 & 0.04 & 0.07 & 0.04 & 0.10 & 0.04 & 0.09 & 0.04 \\
ball\_in\_cup-catch & 810.11 & 28.83 & 898.36 & 15.60 & 924.68 & 12.56 & 940.92 & 8.41 & 943.71 & 10.19 & 946.41 & 8.14 \\
cartpole-balance\_sparse & 254.44 & 25.05 & 327.53 & 29.46 & 368.55 & 32.03 & 382.34 & 33.94 & 397.54 & 36.05 & 430.69 & 36.68 \\
cartpole-swingup & 525.04 & 9.08 & 589.20 & 8.65 & 618.13 & 9.27 & 646.70 & 9.64 & 673.50 & 12.70 & 696.68 & 11.88 \\
finger-spin & 123.71 & 5.88 & 174.48 & 7.20 & 201.30 & 8.27 & 217.71 & 8.55 & 217.13 & 9.70 & 230.65 & 9.63 \\
reacher-hard & 478.71 & 19.47 & 723.58 & 15.48 & 806.71 & 13.85 & 849.64 & 12.68 & 879.62 & 9.41 & 891.60 & 10.64 \\
cheetah-run & 379.54 & 12.04 & 511.85 & 13.00 & 563.08 & 10.13 & 600.29 & 11.29 & 621.27 & 10.00 & 642.45 & 10.89 \\
walker-walk & 287.32 & 8.45 & 321.56 & 8.09 & 330.85 & 8.80 & 335.09 & 8.16 & 344.64 & 8.55 & 348.75 & 8.29 \\
walker-run & 122.34 & 4.21 & 148.59 & 5.14 & 162.94 & 6.41 & 173.03 & 5.40 & 174.46 & 6.84 & 184.43 & 5.45 \\
\bottomrule
\end{tabular}
\label{cma-full-results-a}
\end{table}

\begin{table}[h]
\caption{Full results for CMA-ES on continuous control with limited interaction: mean plus-minus two standard errors.}
\centering
\begin{tabular}{l S[table-format=3.2] @{${}\pm{}$} S[table-format=2.2] S[table-format=3.2] @{${}\pm{}$} S[table-format=2.2] S[table-format=3.2] @{${}\pm{}$} S[table-format=2.2] S[table-format=3.2] @{${}\pm{}$} S[table-format=2.2] } 
\toprule
Environment & \multicolumn{2}{c}{3,5K} & \multicolumn{2}{c}{4K} & \multicolumn{2}{c}{4,5K} & \multicolumn{2}{c}{5K} \\
\midrule
2d-navigation-circles & 0.50 & 0.02 & 0.50 & 0.02 & 0.51 & 0.02 & 0.54 & 0.02 \\
2d-navigation-boxes & 0.20 & 0.03 & 0.20 & 0.02 & 0.26 & 0.03 & 0.24 & 0.03 \\
2d-reacher-fifteen-poles & 0.12 & 0.05 & 0.10 & 0.04 & 0.06 & 0.03 & 0.06 & 0.03 \\
2d-reacher-thirty-poles & 0.09 & 0.04 & 0.14 & 0.05 & 0.06 & 0.03 & 0.12 & 0.05 \\
ball\_in\_cup-catch & 952.83 & 7.85 & 954.25 & 8.73 & 959.55 & 6.27 & 960.81 & 6.76 \\
cartpole-balance\_sparse & 405.10 & 34.60 & 437.27 & 38.91 & 460.61 & 40.94 & 448.40 & 38.79 \\
cartpole-swingup & 695.48 & 12.94 & 713.88 & 13.50 & 726.56 & 13.12 & 733.88 & 13.89 \\
finger-spin & 231.54 & 9.04 & 235.70 & 7.97 & 242.29 & 10.69 & 234.76 & 8.77 \\
reacher-hard & 899.25 & 10.09 & 908.92 & 8.84 & 905.75 & 11.17 & 918.26 & 8.74 \\
cheetah-run & 650.73 & 10.05 & 657.26 & 9.82 & 678.48 & 10.52 & 675.30 & 13.50 \\
walker-walk & 348.28 & 8.80 & 353.03 & 8.74 & 346.99 & 8.01 & 360.31 & 9.79 \\
walker-run & 184.04 & 6.34 & 186.70 & 6.01 & 188.03 & 8.48 & 181.76 & 7.76 \\
\bottomrule
\end{tabular}
\label{cma-full-results-b}
\end{table}

\pagebreak

\subsection{MCTS-ABS}
\begin{table}[h]
\caption{Full results for MCTS-ABS on continuous control with limited interaction: mean plus-minus two standard errors.}
\centering
\begin{tabular}{l S[table-format=3.2] @{${}\pm{}$} S[table-format=2.2] S[table-format=3.2] @{${}\pm{}$} S[table-format=2.2] S[table-format=3.2] @{${}\pm{}$} S[table-format=2.2] S[table-format=3.2] @{${}\pm{}$} S[table-format=2.2] S[table-format=3.2] @{${}\pm{}$} S[table-format=2.2] S[table-format=3.2] @{${}\pm{}$} S[table-format=2.2] } 
\toprule
Environment & \multicolumn{2}{c}{0,5K} & \multicolumn{2}{c}{1K} & \multicolumn{2}{c}{1,5K} & \multicolumn{2}{c}{2K} & \multicolumn{2}{c}{2,5K} & \multicolumn{2}{c}{3K} \\
\midrule
2d-navigation-circles & 0.96 & 0.01 & 0.97 & 0.00 & 0.97 & 0.00 & 0.97 & 0.00 & 0.97 & 0.00 & 0.97 & 0.00 \\
2d-navigation-boxes & 0.32 & 0.04 & 0.52 & 0.04 & 0.62 & 0.04 & 0.65 & 0.04 & 0.73 & 0.03 & 0.74 & 0.03 \\
2d-reacher-fifteen-poles & 0.14 & 0.05 & 0.19 & 0.06 & 0.22 & 0.06 & 0.22 & 0.06 & 0.25 & 0.06 & 0.22 & 0.06 \\
2d-reacher-thirty-poles & 0.16 & 0.05 & 0.28 & 0.06 & 0.25 & 0.06 & 0.25 & 0.06 & 0.22 & 0.06 & 0.23 & 0.06 \\
ball\_in\_cup-catch & 888.98 & 27.04 & 938.40 & 15.49 & 957.22 & 8.54 & 961.29 & 8.12 & 969.31 & 5.02 & 963.12 & 7.97 \\
cartpole-balance\_sparse & 361.23 & 39.64 & 355.61 & 33.88 & 409.53 & 40.44 & 392.47 & 38.81 & 411.01 & 38.73 & 393.84 & 42.31 \\
cartpole-swingup & 494.19 & 12.21 & 494.70 & 11.66 & 494.11 & 10.50 & 496.55 & 9.46 & 493.51 & 9.46 & 495.05 & 9.64 \\
finger-spin & 96.46 & 5.64 & 150.48 & 6.43 & 176.29 & 7.08 & 196.35 & 7.96 & 213.95 & 8.93 & 216.90 & 8.67 \\
reacher-hard & 726.80 & 36.24 & 784.56 & 32.83 & 844.81 & 24.28 & 822.89 & 34.55 & 849.97 & 32.61 & 860.29 & 31.97 \\
cheetah-run & 244.05 & 4.42 & 265.14 & 5.02 & 283.39 & 4.67 & 293.54 & 4.47 & 301.00 & 5.21 & 300.22 & 6.21 \\
walker-walk & 169.25 & 12.78 & 192.59 & 13.76 & 203.42 & 14.54 & 220.94 & 13.78 & 213.56 & 15.10 & 217.90 & 14.66 \\
walker-run & 56.58 & 5.10 & 66.61 & 5.50 & 70.77 & 6.02 & 78.35 & 6.32 & 76.29 & 7.09 & 85.91 & 8.12 \\
\bottomrule
\end{tabular}
\label{mcts_abs-full-results-a}
\end{table}

\begin{table}[h]
\caption{Full results for MCTS-ABS on continuous control with limited interaction: mean plus-minus two standard errors.}
\centering
\begin{tabular}{l S[table-format=3.2] @{${}\pm{}$} S[table-format=2.2] S[table-format=3.2] @{${}\pm{}$} S[table-format=2.2] S[table-format=3.2] @{${}\pm{}$} S[table-format=2.2] S[table-format=3.2] @{${}\pm{}$} S[table-format=2.2] } 
\toprule
Environment & \multicolumn{2}{c}{3,5K} & \multicolumn{2}{c}{4K} & \multicolumn{2}{c}{4,5K} & \multicolumn{2}{c}{5K} \\
\midrule
2d-navigation-circles & 0.97 & 0.00 & 0.97 & 0.00 & 0.97 & 0.00 & 0.97 & 0.00 \\
2d-navigation-boxes & 0.75 & 0.03 & 0.76 & 0.03 & 0.79 & 0.03 & 0.77 & 0.03 \\
2d-reacher-fifteen-poles & 0.23 & 0.06 & 0.31 & 0.07 & 0.25 & 0.06 & 0.21 & 0.06 \\
2d-reacher-thirty-poles & 0.28 & 0.06 & 0.26 & 0.06 & 0.24 & 0.06 & 0.28 & 0.06 \\
ball\_in\_cup-catch & 962.09 & 8.71 & 968.06 & 6.46 & 969.06 & 5.43 & 969.32 & 5.33 \\
cartpole-balance\_sparse & 386.50 & 39.81 & 408.03 & 38.52 & 431.32 & 38.42 & 430.81 & 37.66 \\
cartpole-swingup & 495.99 & 10.69 & 488.70 & 10.67 & 489.97 & 10.57 & 483.32 & 9.30 \\
finger-spin & 220.78 & 12.56 & 234.05 & 12.98 & 243.73 & 13.73 & 250.24 & 12.07 \\
reacher-hard & 861.08 & 32.33 & 859.15 & 34.83 & 873.32 & 31.11 & 867.11 & 37.20 \\
cheetah-run & 303.65 & 5.49 & 310.73 & 5.77 & 316.43 & 5.23 & 322.25 & 5.22 \\
walker-walk & 217.39 & 18.08 & 215.44 & 21.31 & 221.30 & 15.95 & 238.53 & 19.68 \\
walker-run & 91.15 & 8.83 & 94.75 & 10.10 & 87.83 & 9.63 & 90.75 & 10.82 \\
\bottomrule
\end{tabular}
\label{mcts_abs-full-results-b}
\end{table}

\pagebreak

\subsection{VOOT}
\begin{table}[h]
\caption{Full results for VOOT on continuous control with limited interaction: mean plus-minus two standard errors.}
\centering
\begin{tabular}{l S[table-format=3.2] @{${}\pm{}$} S[table-format=2.2] S[table-format=3.2] @{${}\pm{}$} S[table-format=2.2] S[table-format=3.2] @{${}\pm{}$} S[table-format=2.2] S[table-format=3.2] @{${}\pm{}$} S[table-format=2.2] S[table-format=3.2] @{${}\pm{}$} S[table-format=2.2] S[table-format=3.2] @{${}\pm{}$} S[table-format=2.2] } 
\toprule
Environment & \multicolumn{2}{c}{0,5K} & \multicolumn{2}{c}{1K} & \multicolumn{2}{c}{1,5K} & \multicolumn{2}{c}{2K} & \multicolumn{2}{c}{2,5K} & \multicolumn{2}{c}{3K} \\
\midrule
2d-navigation-circles & 0.97 & 0.00 & 0.97 & 0.00 & 0.97 & 0.00 & 0.97 & 0.00 & 0.97 & 0.00 & 0.97 & 0.00 \\
2d-navigation-boxes & 0.43 & 0.04 & 0.64 & 0.04 & 0.74 & 0.03 & 0.78 & 0.03 & 0.84 & 0.02 & 0.85 & 0.02 \\
2d-reacher-fifteen-poles & 0.25 & 0.06 & 0.28 & 0.06 & 0.29 & 0.06 & 0.34 & 0.07 & 0.28 & 0.06 & 0.34 & 0.07 \\
2d-reacher-thirty-poles & 0.33 & 0.07 & 0.31 & 0.07 & 0.39 & 0.07 & 0.35 & 0.07 & 0.38 & 0.07 & 0.41 & 0.07 \\
ball\_in\_cup-catch & 883.12 & 23.24 & 932.01 & 13.16 & 957.04 & 7.92 & 958.58 & 7.39 & 965.63 & 4.78 & 966.69 & 5.09 \\
cartpole-balance\_sparse & 419.35 & 40.62 & 657.16 & 54.32 & 706.29 & 47.18 & 813.97 & 46.62 & 867.83 & 37.61 & 880.61 & 39.70 \\
cartpole-swingup & 574.40 & 14.78 & 618.99 & 12.91 & 642.43 & 12.66 & 634.39 & 13.64 & 663.77 & 13.12 & 652.51 & 14.77 \\
finger-spin & 49.82 & 3.59 & 91.07 & 4.98 & 120.08 & 5.30 & 143.32 & 5.63 & 158.78 & 6.21 & 174.27 & 7.03 \\
reacher-hard & 575.75 & 28.20 & 686.41 & 27.20 & 729.73 & 23.42 & 768.62 & 22.35 & 811.46 & 14.97 & 806.55 & 21.22 \\
cheetah-run & 174.05 & 3.16 & 220.09 & 4.48 & 249.30 & 4.69 & 262.62 & 4.07 & 269.02 & 4.58 & 274.79 & 4.21 \\
walker-walk & 168.57 & 9.30 & 225.02 & 7.79 & 245.29 & 6.91 & 248.63 & 8.13 & 247.43 & 8.60 & 248.16 & 8.41 \\
walker-run & 59.16 & 3.34 & 73.79 & 4.84 & 83.90 & 5.28 & 90.45 & 5.49 & 99.38 & 4.82 & 95.13 & 5.85 \\
\bottomrule
\end{tabular}
\label{voot-full-results-a}
\end{table}

\begin{table}[h]
\caption{Full results for VOOT on continuous control with limited interaction: mean plus-minus two standard errors.}
\centering
\begin{tabular}{l S[table-format=3.2] @{${}\pm{}$} S[table-format=2.2] S[table-format=3.2] @{${}\pm{}$} S[table-format=2.2] S[table-format=3.2] @{${}\pm{}$} S[table-format=2.2] S[table-format=3.2] @{${}\pm{}$} S[table-format=2.2] } 
\toprule
Environment & \multicolumn{2}{c}{3,5K} & \multicolumn{2}{c}{4K} & \multicolumn{2}{c}{4,5K} & \multicolumn{2}{c}{5K} \\
\midrule
2d-navigation-circles & 0.97 & 0.00 & 0.97 & 0.00 & 0.97 & 0.00 & 0.97 & 0.00 \\
2d-navigation-boxes & 0.86 & 0.02 & 0.88 & 0.02 & 0.89 & 0.02 & 0.88 & 0.02 \\
2d-reacher-fifteen-poles & 0.34 & 0.07 & 0.35 & 0.07 & 0.31 & 0.07 & 0.28 & 0.06 \\
2d-reacher-thirty-poles & 0.40 & 0.07 & 0.40 & 0.07 & 0.34 & 0.07 & 0.37 & 0.07 \\
ball\_in\_cup-catch & 967.60 & 5.57 & 968.89 & 5.55 & 968.17 & 5.30 & 971.11 & 4.34 \\
cartpole-balance\_sparse & 921.36 & 33.39 & 944.02 & 24.26 & 896.34 & 36.49 & 927.52 & 31.98 \\
cartpole-swingup & 646.70 & 13.96 & 665.42 & 11.69 & 662.59 & 13.00 & 654.54 & 13.28 \\
finger-spin & 177.01 & 7.34 & 191.27 & 7.64 & 188.10 & 7.36 & 190.72 & 7.60 \\
reacher-hard & 823.39 & 17.47 & 830.63 & 18.09 & 844.25 & 14.50 & 848.15 & 12.73 \\
cheetah-run & 284.04 & 4.76 & 286.00 & 5.94 & 293.92 & 5.15 & 293.34 & 4.89 \\
walker-walk & 249.27 & 8.87 & 258.36 & 7.07 & 257.15 & 8.27 & 259.83 & 6.71 \\
walker-run & 100.57 & 5.33 & 101.23 & 5.13 & 104.59 & 5.38 & 105.14 & 5.35 \\
\bottomrule
\end{tabular}
\label{voot-full-results-b}
\end{table}

\pagebreak

\end{landscape}

\end{document}